\documentclass[runningheads]{llncs}
\usepackage{graphicx}
\usepackage{amsmath,amssymb} % define this before the line numbering.
\usepackage{color}
\usepackage{graphicx}
\usepackage{amsmath,amssymb} % define this before the line numbering.
\usepackage[width=122mm,left=12mm,paperwidth=146mm,height=193mm,top=12mm,paperheight=217mm]{geometry}

\usepackage{graphicx}
\usepackage{array}
\usepackage{multirow}
\usepackage{url}
\usepackage[title]{appendix}

\begin{document}
\title{Long-Term Cloth-Changing Person Re-identification} 
% Replace with your title
\makeatletter
\def\blfootnote{\xdef\@thefnmark{}\@footnotetext}
\makeatother
\titlerunning{Long-Term Cloth-Changing Person Re-identification}
% Replace with a meaningful short version of your title
%
%Xuelin Qian (Fudan University) <xuelinq92@gmail.com>
%Yanwei Fu (Fudan Univ.) <yanweifu@fudan.edu.cn>
%Tao Xiang (Queen Mary, University of London, UK) <t.xiang@qmul.ac.uk>
%Wenxuan Wang (Fudan University) <wxwang.iris@gmail.com>
%Jie Qiu (Nara Institute of Science and Technology) <qiu.jie.qf3@is.naist.jp>
%Yang Wu (Nara Institute of Science and Technology) <yangwu@rsc.naist.jp>
%Yu-Gang Jiang (Fudan University) <ygj@fudan.edu.cn>
%Xiangyang Xue (Fudan University) <xyxue@fudan.edu.cn>

%\orcidID{}
\author{Xuelin Qian$^{1}$ \and 
~Wenxuan Wang$^{1}$ \and
~Li Zhang$^{1}$ \and
~Fangrui Zhu$^{1}$ \and
~Yanwei Fu$^{1}$ \and \\
~Tao Xiang$^{2}$ \and
~Yu-Gang Jiang$^{1}$ \and
~Xiangyang Xue$^{1}$
}

%
%Please write out author names in full in the paper, i.e. full given and family names. 
%If any authors have names that can be parsed into FirstName LastName in multiple ways, please include the correct parsing, in a comment to the volume editors:
%\index{Lastnames, Firstnames}
%(Do not uncomment it, because you may introduce extra index items if you do that, we will use scripts for introducing index entries...)
\authorrunning{X. Qian, W. Wang, L. Zhang, F. Zhu, Y. Fu, T. Xiang, Y. Jiang, X. Xue}
% Replace with shorter version of the author list. If there are more authors than fits a line, please use A. Author et al.
%

\institute{
\textsuperscript{1}Fudan University
\textsuperscript{2}University of Surrey \\
\email{\{xlqian15,wxwang19,lizhangfd,18210980021,yanweifu,ygj,xyxue\}@fudan.edu.cn t.xiang@surrey.ac.uk}
%
% example:
%Princeton University, Princeton NJ 08544, USA \and
%Springer Heidelberg, Tiergartenstr. 17, 69121 Heidelberg, Germany
}

\maketitle              
% typeset the header of the contribution
%
\begin{abstract}
Person re-identification (Re-ID) aims to match a target person across camera views at different locations and times. Existing Re-ID studies focus on the short-term cloth-consistent setting, under which a person re-appears in different camera views with the same outfit. A discriminative feature representation learned by existing deep Re-ID models is thus dominated by the visual appearance of clothing.  In this work, we focus on a much more difficult yet practical setting where person matching is conducted over long-duration, \emph{e.g.}, over days and months and therefore inevitably under the new challenge of changing clothes.  This problem, termed Long-Term Cloth-Changing (LTCC) Re-ID is much understudied due to the lack of large scale datasets. The first contribution of this work is a new LTCC dataset containing people captured over a long period of time with frequent clothing changes. 
As a second contribution, we propose a novel Re-ID method specifically designed to address the cloth-changing challenge. Specifically, we consider that under cloth-changes, soft-biometrics such as body shape would be more reliable. We, therefore, introduce a shape embedding module as well as a cloth-elimination shape-distillation module aiming to eliminate the now unreliable clothing appearance features and focus on the body shape information. 
Extensive experiments show that superior performance is achieved by the proposed model on the new LTCC dataset. The dataset is available on the project website: \url{https://naiq.github.io/LTCC_Perosn_ReID.html}.
\end{abstract}

%===========================================================
\section{Introduction}

%Re-ID
Person re-identification (Re-ID) aims at identifying and associating a person at different locations and times monitored by a distributed camera network.
It underpins many crucial applications such as multi-camera tracking \cite{wang2008correspondencefree}, crowd counting \cite{chan2009counting}, and multi-camera activity analysis \cite{berclaz2008behavioral_maps}.

\begin{figure}
\begin{centering}
\includegraphics[scale=0.35]{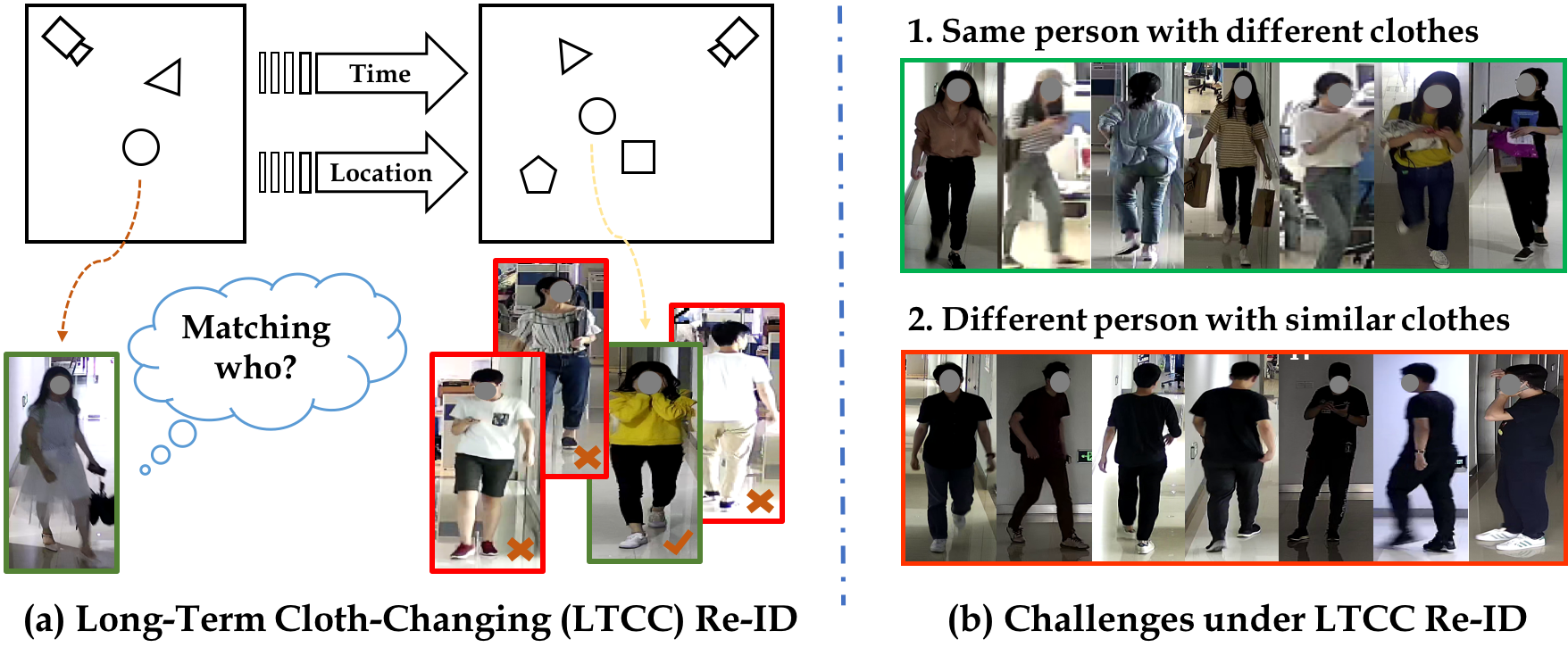} 
\caption{Illustration of the long-term cloth-changing Re-ID task and dataset. 
The task is to match the same person under cloth-changes from different views, and the dataset contains same identities with diverse clothes.} 
\label{fig:intro}
\end{centering}
\end{figure}

% Existing dataset
Re-ID is inherently challenging as a person's appearance often undergoes dramatic changes across camera views.
Such a challenge is captured in existing Re-ID datasets, including Market-1501 \cite{market1501}, DukeMTMC \cite{zheng2017unlabeled} and CUHK03 \cite{deepreid}, where the appearance changes are caused by changes in 
body pose,
illumination~\cite{BakCL18}, occlusion~\cite{zheng2015partial,miao2019pose,he2019foreground} and camera view angle~\cite{qian2017pose,sun2019dissecting}.
However, it is noted that in these popular datasets each  person is captured across different camera views within a short period of time on the same day. As result, each wears the same outfit.  
We call the Re-ID under this setting as short-term cloth-consistent (STCC) Re-ID. This setting has a profound impact on the design of existing Re-ID models -- most existing models are based on deep neural networks (DNNs) that extract feature representations invariant to these changes. Since faces are of too low resolutions to be useful, these representations are dominated by the clothing appearance. 

% new datasets
In this paper, we focus on a more challenging yet practical new Long-Term Cloth-Changing (LTCC) setting (see Fig.~\ref{fig:intro} (a)). Under this setting, a person is matched over a much longer period of time, \textit{e.g.}, days or even months. As a result, clothing changes are commonplace, though different parts of the outfit undergo changes at different frequencies -- the top is more often than the bottom and shoe changes are the least common.  Such a setting is not only realistic but also crucial for certain applications. For instance, if the objective is to capture a criminal suspect after a crime is committed, he/she may change outfits even over short-term as a disguise. However, LTCC Re-ID is largely ignored so far. One primary reason is the lack of large-scale datasets featuring clothing changes. 

The first contribution of this paper is thus to provide such a dataset to fill in the gap between research and real-world applications. The new Long-Term Cloth-Changing (LTCC) dataset (see Fig.~\ref{fig:intro} (b)) is collected over two months. 
It contains  $17,138$ images of $152$ identities with $478$ different outfits captured from $12$ camera views. 
This dataset was completed under pedestrian detection and careful manual labeling of both person identities and clothing outfit identities. Due to the long duration of data collection, it also features drastic illumination, viewing angle, and pose changes as in existing STCC datasets, but additionally clothing and carrying changes and occasionally hairstyle changes. Furthermore, it also includes many persons wearing similar clothing, as shown in Fig.\ref{fig:intro} (b). 

% how to do, use body shape, keypoint
With cloth-changing now commonplace in LTCC Re-ID, existing Re-ID models are expected to struggle (see Tab. \ref{tab:Results-LTCC}) because they assume that the clothing appearance is consistent and relies on clothing features to distinguish people from each other. Instead, it is now necessary to explore identity-relevant biological traits (\textit{i.e.}, soft-biometrics) rather than cloth-sensitive appearance features. A number of naive approaches can be considered. First, can a DNN `does the magic' again, \textit{i.e.}, figures out automatically what information is cloth-change-invariant? The answer is no, because a) the soft-biometrics information is subtle and hard to compute without any network architectural change to assist in the extraction; and b) in a real-world, some people will wear the same outfit or at least keep part of the outfit unchanged (\textit{e.g.}, shoes) even over a long duration. The network thus receives a mixed signal regarding what information is discriminative. Second, would adding a cloth-changing detector be a solution so that models can be switched accordingly? The answer is also negative since detecting changes for the same person needs person Re-ID to be solved in the first place.     

To overcome these difficulties, we propose a novel DNN for LTCC Re-ID. The key idea is to remove the cloth-appearance related information completely and only focus on view/pose-change-insensitive body shape information. To this end, we introduce a Shape Embedding (SE) to help shape feature extraction and a Cloth-Elimination Shape-Distillation (CESD) module to eliminate cloth-related information.
Concretely, the SE module aims to extract body pose features. This is achieved by encoding the position/semantic information of human body joints, and leveraging the relation network~\cite{santoro2017simple} to explore the implicit correlations between each pair of body joints. 
The CESD module, on the other hand, is designed to learn identity-relevant biological representation.
Based on the features extracted from SE module, we adaptively distill the shape information by re-scaling the original image feature.
To better disentangle the identity-relevant features from the residual (\textit{i.e.}, the difference between the original information and the re-scaled shape information), we explicitly add the clothing identity constrain to ensure the identity-irrelevant clothing feature to be eliminated.

% contribution
\noindent \textbf{Contribution.} 
Our contributions are as follows:
(1) We introduce a new Long-Term Cloth-Changing (LTCC) dataset, designed to study the more challenging yet practical LTCC Re-ID problem. %The proposed dataset is collected in natural surveillance system without any human intervention, and it contains identities with cloth-changing under multi-cameras.
(2) We propose a novel model for LTCC Re-ID. It contains a  shape embedding module that efficiently extracts discriminative biological structural feature from keypoints, and a cloth-elimination shape-distillation module that learns to disentangle identity-relevant features from the cloth-appearance features.
(3) Extensive experiments validate the efficacy of our Re-ID models in comparison with existing Re-ID models.

% %%%%%%%%%%%%%%%%%%%%%%%%%%%%%%%%%%%%
% ------------------------------------
%              Related Work
% ------------------------------------
% %%%%%%%%%%%%%%%%%%%%%%%%%%%%%%%%%%%%
\section{Related Work}

\noindent \textbf{Short-Term Cloth-Consistent Re-ID. } 
With the popularization of surveillance system in the real-world, person re-identification task attracts more and more attention. As mentioned earlier, almost all existing Re-ID datasets \cite{wei2018person,ristani2016MTMC,market1501} were captured during short-period of time. As a result,  for the same person, the clothing appearances are more or less consistent. 
In the deep-learning era, more efforts have been made in developing approaches for automatic person re-identification by learning discriminative features \cite{wang2019color,zheng2019joint} or robust distance metrics \cite{paisitkriangkrai2015learning,shen2018end}.
These models are robust against changes caused by pose, illumination and view angles as they are plenty in those datasets. However, they are vulnerable to clothing changes as the models are heavily reliant on the clothing appearance consistency. 
%, and the number of cameras, and spatial size of the environment and people may be unknown. 

\noindent \textbf{Long-Term Cloth-Changing Re-ID Datasets. } 
Barbosa \textit{et al.} \cite{barbosa2012re} proposed the first cloth-changing Re-ID dataset. However, the dataset is too small to be useful for deep learning based Re-ID. More recently iQIYI-VID \cite{liu2018iqiyi} is introduced which is not purposefully built for LTCC but does contain some cloth-changing images.  However, it is extracted from on-line videos, unrepresentative of real-world scenarios and lack of challenges caused illumination change and occlusion. Similarly, Huang \textit{et al.} \cite{huang2019celebrities,huang2019beyond} collect Celebrities-ReID dataset containing clothing variations. Nevertheless, the celebrity images are captured in high quality by professional photographers, so unsuited for the main application of Re-ID, i.e. video surveillance using CCTV cameras. 
Another related work/dataset is Re-ID in photo albums \cite{joon2015person,zhang2015beyond}.
It involves person detection and recognition with high-resolution images where face is the important clue to solve this task. 
However, this setting is not consistent with canonical Person Re-ID, where pedestrians are captured by non-overlapped surveillances with low resolution.
Note that a concurrent work~\cite{yang2019person} also introduced an LTCC dataset, called PRCC which is still not released. 
Though featuring outfit changes, PRCC is a short-term cloth-changing dataset so it contains less drastic clothing changes and bare hairstyle changes. Further, with only 3 cameras instead of 12 in our LTCC dataset, it is limited in view-angle and illumination changes.  In contrast, our LTCC Re-ID aims at matching persons over long time from more cameras, with more variations in visual appearance, \textit{e.g.}, holding different bags or cellphones, and wearing hats as shown in Fig.~\ref{fig:intro}(b). %To this end, the proposed LTCC dataset collected in a natural and long-term way, which contains abundant variations in aspects of clothes, poses, occlusion and lighting conditions, etc..

\noindent \textbf{Long-Term Cloth-Changing Re-ID Models.}  Recently, Xue \textit{et al.} \cite{xue2018clothing} particularly address the  cloth-change challenge by downplaying clothing information and emphasizing face. In our dataset, face resolution is too low to be useful and we take a more extreme approach to remove clothing information completely.  Zheng \textit{et al.} \cite{zheng2019joint} propose to switch the appearance or structure codes and leverage the generated data to improve the learned Re-ID features. Yang \textit{et al.} \cite{yang2019person} introduce a method to capture the contour sketch representing the body features. These two works mainly focus on learning invariant and discriminative Re-ID features by taking clothing information into account. In contrast, our model focuses solely on extracting soft-biometrics features and removes the model dependency on clothing information completely.

\begin{figure}
\begin{centering}
\includegraphics[scale=0.4]{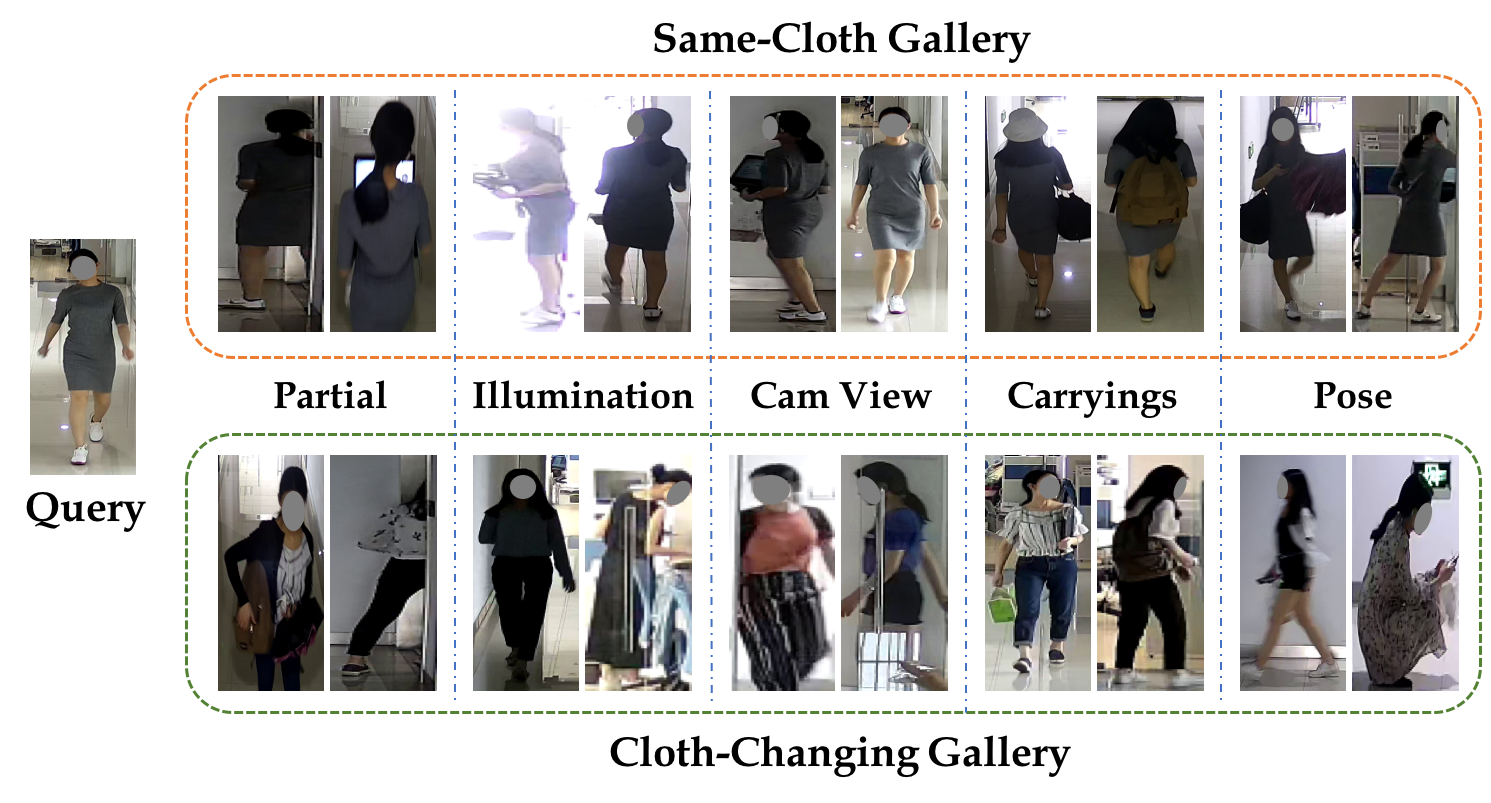} 
\caption{Examples of one person wearing the same and different clothes in LTCC dataset. There exist various illumination, occlusion, camera view, carrying and pose changes. \label{fig:LTCC} }
\end{centering}
\end{figure}

% %%%%%%%%%%%%%%%%%%%%%%%%%%%%%%%%%%%%
% ------------------------------------
%              Dataset
% ------------------------------------
% %%%%%%%%%%%%%%%%%%%%%%%%%%%%%%%%%%%%
\section{Long-Term Cloth-Changing (LTCC) Dataset \label{new_dataset}} 

\noindent \textbf{Data Collection. }
To facilitate the study of LTCC Re-ID, we collect a new Long-Term Cloth-Changing (LTCC) person Re-ID dataset. Different from previous datasets \cite{market1501,cuhk01,zheng2017unlabeled,deepreid}, this dataset aims to support the research of long-term person re-identification with the added challenge of cloth changes. During dataset collection, an existing CCTV network is utilized which is composed of twelve cameras installed on three floors in an office building. A total of  24-hour videos are captured over two months. Person images are then obtained by applying the Mask-RCNN \cite{he2017mask} detector.

\noindent \textbf{Statistics. }
We release the first version of LTCC dataset with this paper, which contains 17,138 person images of 152 identities, as shown in Fig. \ref{fig:LTCC}. 
Each identity is captured by at least two cameras. 
To further explore the cloth-changing Re-ID scenario, we assume that different people will not wear identical outfits (however visually similar they may be), and annotate each image with a cloth label as well. Note that the changes of the hairstyle or carrying items, \textit{e.g.}, hat, bag or laptop, do not affect the cloth label. 
Finally, dependent on whether there is a cloth-change, the dataset can be divided into two subsets: one cloth-change set where $91$ persons appearing with $417$ different sets of outfits in $14,756$ images, and one cloth-consistent subset containing the remaining $61$ identities with $2,382$ images without outfit changes. On average, there are $5$ different clothes for each cloth-changing person, with the numbers of outfit changes ranging from 2 to 14. 

\begin{figure}
\begin{centering}
\includegraphics[scale=0.45]{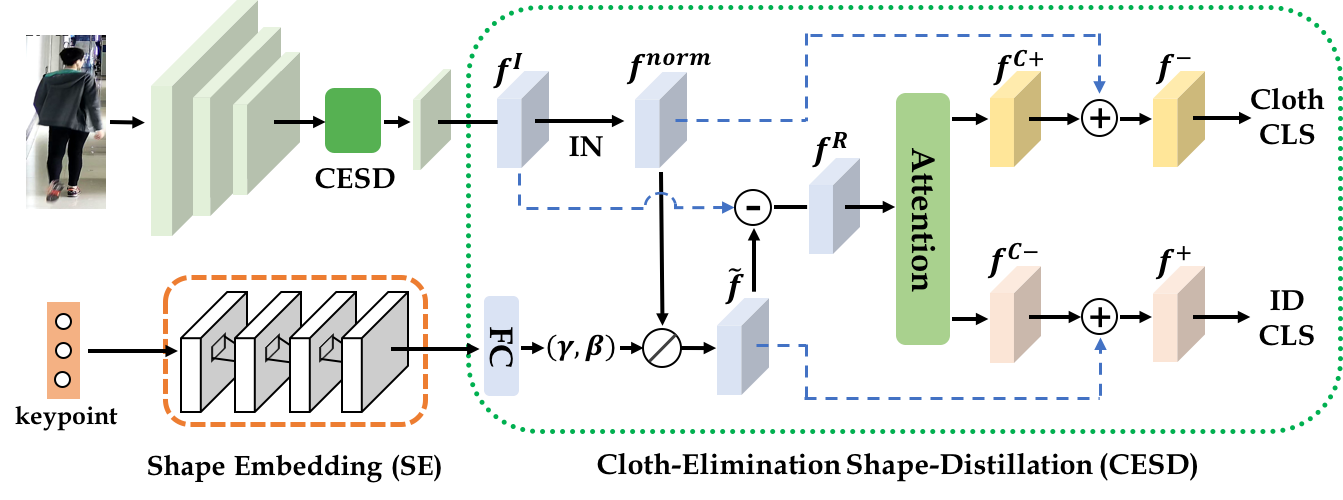} 
\caption{Illustration of our framework and the details of Cloth-Elimination Shape-Distillation (CESD) module. Here, we introduce Shape Embedding (SE) module to extract structural features from human keypoints, followed by learning identity-sensitive and cloth-insensitive representations using the CESD module. 
There are two CESD modules (green solid box and green dash box). Both share the same SE module, but we only show the detail of CESD in the second one. `$\oslash$' denotes the operation of re-scale in the Eq. \ref{eq:re-scale}.
\label{fig:overall}}
\end{centering}
\end{figure}

\noindent \textbf{Comparison with Previous Datasets. }
Comparing to the most widely-used cloth-consistent Re-ID datasets such  \cite{market1501,deepreid,zheng2017unlabeled}, our dataset is gathered primarily to address the new LTCC Re-ID task, which challenges the Re-ID  under a more realistic yet difficult setting. Comparing to few cloth-changing datasets \cite{munaro2014one,huang2019celebrities}, ours is collected in a natural and long-term way without any human intervention. Concretely, our dataset includes not only various cloth-changing (\textit{e.g.}, top, bottom, shoe-wear), but also diverse human pose (\textit{e.g.}, squat, run, jump, push, bow), large changes of illumination (\textit{e.g.}, from day to night, indoor lighting) and large variations of occlusion (\textit{e.g.}, self-occlusion, partial), as shown in Fig.~\ref{fig:LTCC}. More details can be found in Supplementary Material.

% %%%%%%%%%%%%%%%%%%%%%%%%%%%%%%%%%%%%
% ------------------------------------
%              Methodology
% ------------------------------------
% %%%%%%%%%%%%%%%%%%%%%%%%%%%%%%%%%%%%

\section{Methodology} \label{sec:method}

Under the LTCC Re-ID setting, the clothing appearance becomes unreliable and can be considered as a distraction for Re-ID. We thus aim  to learn to extract  biological traits related to soft biometrics, with a particular focus on  body shape information in this work. Unfortunately, learning the identity-sensitive feature directly from body shape \cite{chao2019gaitset} is a challenging problem on its own. Considering the recent success in  body parsing or human body pose estimation \cite{liang2018look} from RGB images, we propose to extract identity-discriminative shape information whilst eliminating clothing information with the help of an off-the-shelf body pose estimation model. 
% Unfortunately, learning shapes, \textit{e.g.}, pose or gait estimation, is also a high-level task, whose results even produced by deep models~\cite{liang2018look}, may be still likely to be sensitive to many factors, \textit{e.g.}, body occlusion, or illumination conditions. 
Specifically, given a generic DNN backbone, we first introduce a Shape Embedding (SE) module to encode shape information from human body keypoints. Then we propose a Cloth-Elimination Shape-Distillation (CESD) module, which is able to utilize shape embedding to adaptively distill the identity-relevant shape feature and explicitly disentangle the identity-irrelevant clothing information.

\subsection{Shape Embedding (SE)} \label{shape_embedding}
Humans can easily recognize an old friend in an unseen outfit from the back (no face). We conjecture that this is  because the body shape information is discriminative enough for person identification.
Here, `shape' is a general term referring to several \textit{unique} biological traits, \textit{e.g.}, stature and body-part proportion. One intuitive way to represent body shape is to employ joints of human body and model the relations between each pair of points. For example, the relation between points of left-shoulder and left-hip reflects the height of the upper body. 
%However, these features are \textit{implicit} and \textit{fragile} in images. 
%Inspired by adaptive instance normalization \cite{huang2017arbitrary}, we propose a shape embedding module to encode body shape from joints.

\begin{figure}
\begin{centering}
\includegraphics[scale=0.37]{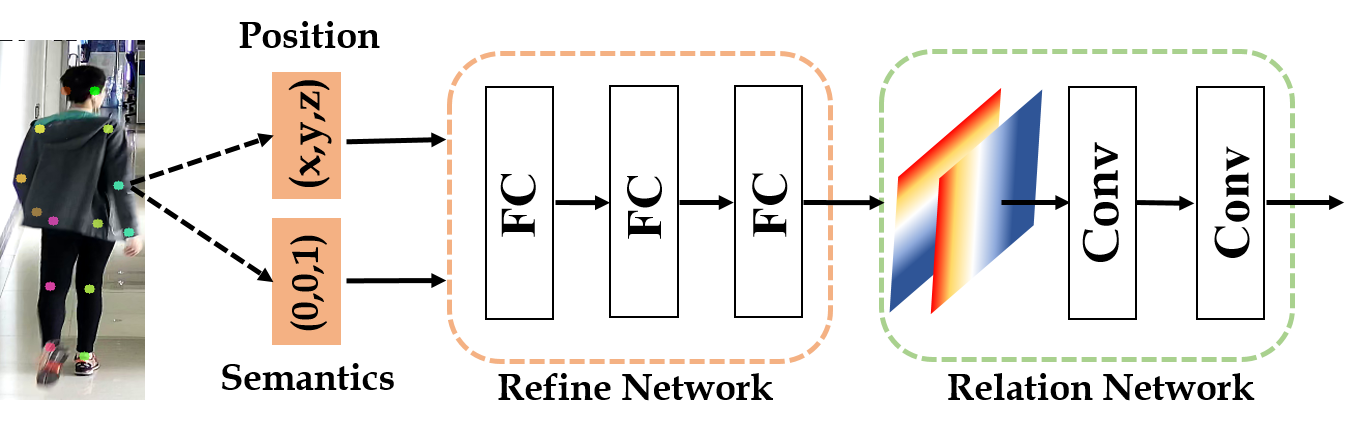} 
\caption{The detailed structure of the Shape Embedding (SE) module. %With the position and semantic keypoint information as the input, we use the refine network to embed them and apply relation network for mining the relation between keypoints for biological human shape information.
\label{fig:SE}}
\end{centering}
\end{figure}

\noindent \textbf{\textcolor{black}{Representation}.} We employ an off-the-shelf pose detector \cite{fang2017rmpe,xiu2018poseflow}, which is trained without using any re-id benchmark data, to detect and localize $n$ human body joints as shown in Fig. \ref{fig:SE}. Each point $n_{i}$ is represented by two attributes, position $P_{i}$ and semantics $S_{i}$ (\textit{i.e.}, corresponding to which body joint). Concretely, $P_{i}=(\frac{x_{i}}{w}, \frac{y_{i}}{h}, \frac{w}{h})$ where $(x_{i}, y_{i})$ denotes the coordinates of joint $i$ in the person image and $(w, h)$ represents the original width and height of the image. $S_{i}$ is a $n$-dimensional one-hot vector to index the keypoint $i$ (\textit{e.g.}, head and knee). If one of the points is undetectable, we set $P_{i}$ as $(-1, -1, \frac{w}{h})$.

\noindent \textbf{\textcolor{black}{Embedding}.} The two keypoint representation parts are first encoded with learnable embedding weights $W$ individually. Then, we employ a refinement network to integrate the two parts and improve the representation of each keypoint, which can be formulated as 
\begin{equation}
f_{i} = \mathcal{F}\left(W_{p}{P_{i}}^\mathsf{T}+W_{s}{S_{i}}^\mathsf{T}\right) \in \mathbb{R}^{d_{2}},
\end{equation}
\noindent where $W_{p}\in\mathbb{R}^{d_1\times 3}$ and $W_{s}\in\mathbb{R}^{d_1\times 13}$ are two different embedding weights; $\mathcal{F}\left(\cdot\right)$ denotes the refinement network with several fully-connected layers to increase the dimension from $d_{1}$ to $d_{2}$. In this paper, we have two hidden layers and set $d_{1}=128$, $d_{2}=2048$.

After the embedding, we now obtain a set of keypoint features $f\in\mathbb{R}^{n\times d_{2}}$. Intuitively, the information of body proportion cannot be captured easily by the feature of a single joint. We still need to represent the relation between each pair of keypoints. To this end, we propose to leverage a relation network for exploiting relations between every two points. As illustrated in Fig.~\ref{fig:SE}, our relation network concatenates features of two different keypoints from all combinations and feeds them into two convolution layers for relation reasoning.
% a layer of Multilayer Perceptron (MLP) for reasoning. 
\textcolor{black}{The final shape embedding feature of $f^{P}$ is obtained by maximizing\footnote{Max pooling is found to be more effective than alternatives such as avg pooling; possible reason is that it is more robust against body pose undergoing dramatic changes.} over all outputs.} 
The whole formulations can be expressed as follows,
\begin{equation}
H_{ij}=[f_{i}~;~f_{j}], ~~~~~f^{p} = \mathrm{GMP}\left(Conv_{\theta}(H)\right)
% f^{P} = \mathrm{GMP}\left(H_{ij}\right), ~~~~~~ H_{ij} = \mathrm{Conv}_{\theta}\left( [f_{i}~;~f_{j}] \right), ~ i,j\in[1,2,\dots,n],
\end{equation}
\noindent where $f^{P}\in\mathbb{R}^{d_{2}}$ and $H\in\mathbb{R}^{n\times n \times d_{2}}$; $\mathrm{GMP}$ and $\theta$ denote global max pooling and parameters of convolution layers, respectively. $[*;*]$ denotes the operation of concatenation.

\subsection{Cloth-Elimination Shape-Distillation (CESD)} \label{CESD}
Person images contain clothing and shape information. Disentangling the two and focusing on the latter is the main idea behind our model for LTCC Re-ID.  Recently, many works on visual reasoning \cite{perez2017learning,perez2018film} or style transfer \cite{huang2017arbitrary,ghiasi2017exploring} are proposed to solve a related similar problem with adaptive normalization. Inspired by these works,  
% we introduce the cloth-elimination shape-distillation module, which is parameterised using keypoint embedding, to remove cloth clues from the input and distill task-specific discriminative features. 
we introduce a cloth-elimination shape-distillation module. \textcolor{black}{It is designed to utilize the shape embedding to explicitly disentangle the clothing and discriminative shape information from person images\footnote{\textcolor{black}{We found empirically that directly using shape embeddings as Re-ID features leads to worse performance. A likely reason is that the detected 2D keypoints may be unreliable due to occlusion. So, we treat them as intermediate ancillary features.}}.
The final extracted feature thus contains the information transferred from shape embedding and the ones from a task-driven attention mechanism.} 
Specifically, we denote both the input image and shape features by $f^{I}\in\mathbb{R}^{h\times w\times c}$ and $f^{P}\in\mathbb{R}^{d_{2}}$, where $h$, $w$, $c$ indicate the height, width and number of channel, respectively.

\noindent \textbf{Shape Distillation.} Inspired by adaptive instance normalization \cite{huang2017arbitrary}, we first try to reduce the original style information
% correlated mostly to the illumination condition, 
in the input by performing instance normalization \cite{ulyanov2016instance}.  Then, we distill the shape information by re-scaling the normalized feature with the parameters $\gamma$ and $\beta$ calculated from $f^{P}$. These two steps can be expressed as,
\begin{equation}
f^{norm} = \frac{f^{I} - \mathrm{E}\left[f^{I}\right]}{\sqrt{\mathrm{Var}\left[f^{I}\right] - \epsilon}},
\end{equation}
\begin{equation}
\begin{aligned}
\widetilde{f} = & \left(1+\mathrm{\Delta}\gamma\right) f^{norm} + \beta, \\
&where~\mathrm{\Delta}\gamma = \mathcal{G}_{s}\left(f^{P}\right), ~~ \beta = \mathcal{G}_{b}\left(f^{P}\right)
\label{eq:re-scale}
\end{aligned}
\end{equation}
\noindent where $\mathrm{E}\left[\cdot\right]$ and $\mathrm{Var}\left[\cdot\right]$ denote the mean and variance of image features calculated per-dimension separately for each sample; $\mathcal{G}_{s}\left(\cdot\right)$ and $\mathcal{G}_{b}\left(\cdot\right)$ are both one fully-connected layer to learn new parameters of scale and bias. Particularly, rather than directly predicting $\gamma$, we output the offset $\mathrm{\Delta}\gamma$ in case of  re-scaling factor value for the feature activation being too low.

\noindent \textbf{Cloth Elimination.} 
% We agree that only estimating shape keypoints is not enough, as it is not as straightforward as appearance, directly encoded in the image data. Consequently, we treat it as an auxiliary and target to purify other discriminative representations from the features that exclude the shape information.
To further enhance the representation of identity-sensitive but cloth-insensitive feature, we propose to eliminate the identity-irrelevant clothing clue from the final image features.
As shown in Fig.~\ref{fig:overall}, given the image feature $f^{I}$ and the transferred feature $\widetilde{f}$, we first obtain the residual feature $f^{R}\in\mathbb{R}^{h\times w\times c}$ by computing the difference between $f^{I}$ and $\widetilde{f}$,
\begin{equation}
f^{R} = f^{I} - \widetilde{f}
\end{equation}
For $f^{R}$, it inevitably includes some discriminative features (\textit{e.g.}, contour) and  features that are sensitive to cloth changes. Since reducing intra-distance of the same identity with different clothing is the primary objective here, we propose to leverage the self-attention mechanism to explicitly disentangle the residual feature into two parts, the cloth-irrelevant feature $f^{C-}\in\mathbb{R}^{h\times w\times c}$ and the cloth-relevant feature $f^{C+}\in\mathbb{R}^{h\times w\times c}$, which can be formulated as follows,
\begin{equation}
\alpha = \phi\left(\mathcal{G}^{2}\left(\mathrm{GAP}\left(f^{R}\right)\right)\right),
\end{equation}
\begin{equation}
f^{C+} = \alpha f^{R}, ~~~ f^{C-} = (1-\alpha) f^{R}
\end{equation} 
\noindent where $\mathcal{G}^{i}$ denotes the $i$-th layer of a convolution neural network with ReLU activation function,  $\mathrm{GAP}$ is the  global average pooling operation, $\phi$ is a sigmoid activation function and $\alpha\in\mathbb{R}^{1\times1\times c}$ is the learned attention weights.

By adding the cloth-irrelevant feature $f^{C-}$ to the re-scaled shape feature $\widetilde{f}$, we add one more convolutional layer to refine it and obtain the identity-relevant feature of $f^{+}$. Analogously, we sum the cloth-relevant feature $f^{C+}$ and the normalized feature $f^{norm}$ followed by a different convolutional layer  to get the final cloth-relevant feature $f^{-}$, that is, 
% By adding the cloth-irrelevant feature $f^{C-}$ to the re-scaled shape feature $\widetilde{f}$, we obtain the identity-releva feature of $f^{+}$, which is more robust to cloth changing. Analogously, we sum over the cloth-relevant feature $f^{C+}$ and the normalized feature $f^{norm}$ to get $f^{-}$. Finally, one convolution layer is followed to refine the output,
\begin{equation}
\label{eq:cloth-elimination1}
f^{+} = \mathrm{Conv}_{\theta}\left(\widetilde{f} + f^{C-}\right), ~~~ f^{-} = \mathrm{Conv}_{\phi}\left(f^{norm} + f^{C+} \right).
\end{equation} 

\subsection{Architecture Details} \label{architecture_details}
Figure~\ref{fig:overall} gives an overview of our framework. Particularly, a widely-used network of ResNet-50 \cite{resnet} is employed as our backbone for image feature embedding. We insert the proposed CESD module after \textit{res3} and \textit{res4} blocks. For each CESD module, we apply two classification losses (person ID and cloth ID) to support the learning of identity-relevant feature $f^{+}$ and cloth-relevant feature $f^{-}$ respectively. Therefore, the overall loss of our framework is, 
\begin{equation}
\label{eq:overall-loss}
\mathcal{L} = \sum^{2}_{i=1}\lambda_{i}\mathcal{L}^{i}_{clothing} + \sum^{2}_{i=1}\mu_{i}\mathcal{L}^{i}_{id},
\end{equation} 
\noindent where $\mathcal{L}^{i}_{clothing}$ and $\mathcal{L}^{i}_{id}$ denote the cross-entropy loss of clothing and identity classification from the $i$-th CESD module, respectively; $\lambda_{i}$ and $\mu_{i}$ are both coefficients which control the contribution of each term. Intuitively, the feature at the deeper layer is more important and more relevant to the task, thus, we empirically set $\lambda_{1}, \lambda_{2}$ to $0.3, 0.6$ and  $\mu_{1}, \mu_{2}$ to $0.5, 1.0$ in all experiments.

% %%%%%%%%%%%%%%%%%%%%%%%%%%%%%%%%%%%%
% ------------------------------------
%              Experiment
% ------------------------------------
% %%%%%%%%%%%%%%%%%%%%%%%%%%%%%%%%%%%%

\section{Experiment} \label{sec:experiment}

\subsection{Experimental Setup}

\begin{table}
\centering{}%
\setlength{\tabcolsep}{0.1mm}{
\begin{tabular}{@{\extracolsep{\fill}}l|c|c|c|c||c|c|c|c}
\hline 
%  & \multicolumn{4}{c||}{Mixed Training} & \multicolumn{4}{c}{Diff-cloth Training}\tabularnewline
\multicolumn{1}{c|}{\multirow{2}{*}{Methods}} & \multicolumn{2}{c|}{Standard} & \multicolumn{2}{c||}{Cloth-changing} & \multicolumn{2}{c|}{Standard$^\dag$} & \multicolumn{2}{c}{Cloth-changing$^\dag$}\tabularnewline
\cline{2-9} 
 & Rank-1 & mAP & Rank-1 & mAP & Rank-1 & mAP & Rank-1 & mAP\tabularnewline
\hline 
LOMO \cite{XQDA} + KISSME \cite{crowduser} & 26.57 & 9.11 & 10.75 & 5.25 & 19.47 & 7.37 & 8.32 & 4.37 \tabularnewline
LOMO \cite{XQDA} + XQDA \cite{XQDA} & 25.35 & 9.54 & 10.95 & 5.56 & 22.52 & 8.21 & 10.55 & 4.95 \tabularnewline
LOMO \cite{XQDA} + NullSpace \cite{NullReid} & 34.83 & 11.92 & 16.45 & 6.29 & 27.59 & 9.43 & 13.37 & 5.34 \tabularnewline
\hline 
% AlexNet \cite{KrizhevskySH12} &  &  &  & \tabularnewline
% VGG19 \cite{simonyan2014very} &  &  &  & \tabularnewline
ResNet-50 (Image) \cite{resnet} & 58.82 & 25.98 & 20.08 & 9.02 & 57.20 & 22.82 & 20.68 & 8.38 \tabularnewline
PCB (Image) \cite{sun2018beyond} & 65.11 & 30.60 & 23.52 & 10.03 & 59.22 & 26.61 & 21.93 & 8.81 \tabularnewline
\hline
HACNN \cite{li2018harmonious} & 60.24 & 26.71 & 21.59 & 9.25 & 57.12 & 23.48 & 20.81 & 8.27 \tabularnewline
MuDeep \cite{qian2019leader} & 61.86 & 27.52 & 23.53 & 10.23 & 56.99 & 24.10 & 18.66 & 8.76\tabularnewline
OSNet \cite{sun2018beyond} & 67.86 & 32.14 & 23.92 & 10.76 & 63.48 & 29.34 & 24.15 & 10.13\tabularnewline
\hline
ResNet-50 (Parsing) \cite{resnet} & 19.87 & 6.64 & 7.51 & 3.75 & 18.86 & 6.16 & 6.28 & 3.46 \tabularnewline
PCB (Parsing) \cite{sun2018beyond} & 27.38 & 9.16 & 9.33 & 4.50 & 25.96 & 7.77 & 10.54 & 4.04
\tabularnewline
ResNet-50 + Face$^\S$ \cite{xue2018clothing} & 60.44 & 25.42 & 22.10 & 9.44 & 55.37 & 22.23 & 20.68 & 8.99
\tabularnewline
\hline
%ResNet-50 (Face) \cite{resnet} &  &  &  &  &  &  &  & \tabularnewline
Ours & \bf 71.39 & \bf34.31 & \bf 26.15 & \bf 12.40 & \bf 66.73 & \bf 31.29 & \bf 25.15 & \bf 11.67 \tabularnewline
\hline 
\end{tabular}}\caption{\label{tab:Results-LTCC} 
Results comparisons of our model and other competitors.
`Standard' and `Cloth-changing' mean the standard setting and cloth-changing setting, respectively. `(Image)' or `(Parsing)' represents that the input data is person image or body parsing.
`$\dag$' denotes that only identities with clothes changing are used for training. `$\S$' indicates our simple implementation of the siamese network using face images detected by \cite{tang2018pyramidbox}.}
\end{table}

\noindent \textbf{Implementation details.} Our model is implemented on the Pytorch framework, and we utilize the weights of ResNet-50 pretrained on ImageNet \cite{deng2009imagenet} for initialization. For key points, we first employ the model from \cite{fang2017rmpe,xiu2018poseflow} for human joints detection to obtain 17 key points. Then, we average 5 points from face (\textit{i.e.}, nose, ear, eye) as one point of `face', given us the  13 human key joints and their corresponding coordinates.  During training, the input images are resized to $384\times192$ and we only apply random erasing \cite{zhong2017random} for data augmentation. SGD is utilized as the optimizer to train networks with mini-batch $32$, momentum $0.9$, and the weight decay factor for L2 regularization is set to $0.0005$. Our model is trained on one NVIDIA TITAN Xp GPU for total 120 epochs with an initial learning rate of $0.01$ and a reduction factor of $0.1$ every 40 epochs. During testing, we concatenate features $f^{+}$ from all CESD modules as the final feature.

\noindent \textbf{Evaluation settings.} We randomly split the LTCC dataset into training and testing sets. The training set consists of $77$ identities, where $46$ people have cloth changes and the rest of $31$ people wear the same outfits during recording. Similarly, the testing set contains $45$ people with changing clothes and $30$ people wearing the same outfits. 
Unless otherwise specified, we use all training samples for training.
For better analyzing the results of long-term cloth-changing Re-ID in detail, we introduce two test settings, as follows,
(1) \textit{Standard Setting}: Following the evaluation in \cite{market1501}, the images in the test set with {the same identity and the same camera view are discarded} when computing evaluation scores, \textit{i.e.}, Rank-k and mAP. In other words, the test set contains both cloth-consistent and cloth-changing examples. 
(2)  \textit{Cloth-changing Setting}: Different from \cite{market1501}, the images with same identity, camera view and clothes are discarded during testing. This setting examines specifically how well a Re-ID model copes with cloth changes.

\subsection{Comparative Results}

\noindent \textbf{LTCC dataset.} We evaluate our proposed method on the LTCC dataset and compare it with several competitors, including hand-crafted features of LOMO \cite{XQDA} + KISSME \cite{crowduser}, LOMO \cite{XQDA} + XQDA \cite{XQDA}, LOMO \cite{XQDA} + NullSpace \cite{NullReid}, deep learning baseline as ResNet-50 \cite{resnet}, PCB \cite{sun2018beyond}, and strong Re-ID methods MuDeep \cite{qian2019leader}, OSNet \cite{sun2018beyond} and HACNN \footnote{\textcolor{black}{OSNet is trained with the size of $384 \times 192$ and the cross-entropy loss as ours for a fair comparison. HACNN is trained with $160 \times 64$ as required by the official code.}} \cite{li2018harmonious}. In addition, we try to leverage human parsing images, which are semantic maps containing body structure information \cite{Gong_2017_CVPR}. 
Please find more quantitative and qualitative studies in Supplementary Material.
% Due to space limitation, please refer to the Supplementaty Material for more experimental resutls on other cloth-changing datasets, \emph{i.e.}, BIWI \cite{munaro2014one} and Celeb-reID \cite{huang2019celebrities}, and more qualitative studies on LTCC dataset.

A number of observations can be made from results in Tab.~\ref{tab:Results-LTCC}. 
(1) Our method beats all competitors with a clear margin under both evaluation settings. As expected, the re-identification results under the cloth-changing setting are much lower than the standard setting, which verify the difficulty of LTCC Re-ID. Under cloth-changing setting, where the training and testing identities both with different clothes, our method surpass the general baseline `ResNet-50 (Image)' with $4.47\%/3.29\%$ of Rank-1/mAP, and it also outperforms strong baseline `PCB (Image)' by $3.22\%/2.86\%$. 
(2) Meanwhile, our model also achieves better results than those advanced Re-ID methods, where some of them are designed with multi-scale or attention mechanism to extract more discriminative features under the general Re-ID problem.
% which shows the existing Re-ID models, targeting at the general Re-ID with cloth-consistency, are fragile to the realistic LTCC Re-ID. 
Such results indicate that our proposed method can better eliminate negative impact of cloth changes and explore more soft-biometrics identity-relevant features.
(3) Following the idea of \cite{xue2018clothing}, we build a siamese network applying the information of face for LTCC Re-ID.
Intuitively, more information, \textit{i.e.,} face, leads to better performance. However, face feature is sensitive to factors of illumination, occlusion and resolution, so it is not the optimal way for the cloth-changing Re-ID.
(4)  Comparing the performance of `ResNet-50 (Parsing)' and `ResNet-50 (Image)', as well as `PCB (Parsing)' and `PCB (Image)', we can find that the models trained on person images have superior accuracy than those using parsing. This suggests that the identity-sensitive structural information is hard to capture directly from images. As a result,  applying it alone for re-identification is unhelpful.

\noindent \textbf{BIWI dataset.} Additional experimental results on BIWI \cite{munaro2014one} dataset are further presented and discussed. Please refer to Supplementary Material for more details about dataset introduction and implementation details. From the results shown in Table \ref{tab:Results-DIWI}, we can make the following observations:
(1) Our method achieves the highest performance on Rank-$1$ at both settings (Rank-1 of 49.51\% and 39.80\% compared to the closest competitor OSNet~\cite{zhou2019omni} which gives 47.14\% and 38.87\% at Still and Walking setting respectively).
(2) Our method beats the state-of-the-art competitor  SPT+ASE~\cite{yang2019person}, which is designed for cloth-changing re-ID specifically, by over 28\% and 21\% on Rank-$1$ under the Still and Walking setting, respectively.
Note that we adopt the same training/test split and pretrain strategy as SPT+ASE~\cite{yang2019person}.
(3) We also notice that the performance of Still setting are generally higher than the Walking setting.
This confirms the conclusion drawn in the main paper as `Still' subset involve fewer variations of pose and occlusions.

\begin{table}
\centering{}%
\setlength{\tabcolsep}{4mm}{
\begin{tabular}{@{\extracolsep{\fill}}l|cc|cc}
\hline 
\multicolumn{1}{c|}{\multirow{2}{*}{Methods}} & \multicolumn{2}{c|}{Still Setting} & \multicolumn{2}{c}{Walking Setting}\tabularnewline
\cline{2-5}
 & Rank-1 & Rank-5 & Rank-1 & Rank-5 \tabularnewline
\hline
ResNet-50 \cite{resnet} & 41.26 & 65.13 & 37.08 & 65.01 \tabularnewline
PCB \cite{sun2018beyond} & 46.36 & 72.37 & 37.89 & \textbf{70.21} \tabularnewline
\hline 
SPT+ASE$^*$ \cite{yang2019person} & 21.31 & 66.10 & 18.66 & 63.88 \tabularnewline
HACNN \cite{li2018harmonious} & 42.71 & 64.89 & 37.32 & 64.27 \tabularnewline
MuDeep \cite{qian2019leader} & 43.46 & \textbf{73.80} & 38.41 & 65.35 \tabularnewline
OSNet \cite{zhou2019omni} & 47.14 & 72.52 & 38.87 & 61.22 \tabularnewline
\hline 
Ours & \textbf{49.51} & 71.44 & \textbf{39.80} & 66.23 \tabularnewline
\hline 
\end{tabular}}
\caption{\label{tab:Results-DIWI} Results of the multi-shot setting on BIWI dataset. `$*$' denotes the results are reported from original paper~\cite{yang2019person}.
Note that, the same baselines (\textit{i.e.}, ResNet-50 \cite{resnet}, PCB~\cite{sun2018beyond}) implemented by us yield better results  than those reported in~\cite{yang2019person}.}
\end{table}

\subsection{Ablation Study}

\begin{table}
\centering{}%
\setlength{\tabcolsep}{1.7mm}{
\begin{tabular}{@{\extracolsep{\fill}}l|ccc|ccc}
\hline 
\multicolumn{1}{c|}{\multirow{2}{*}{Methods}} & \multicolumn{3}{c|}{Standard Setting} & \multicolumn{3}{c}{Cloth-changing Setting}\tabularnewline
\cline{2-7}
 & Rank-1 & Rank-5 & mAP & Rank-1 & Rank-5 & mAP\tabularnewline
\hline
ResNet-50 \cite{resnet} & 57.20 & 71.19 & 22.82 & 20.68 & 31.84 &8.38 \tabularnewline
\hline 
\textcolor{black}{ResNet-50 \cite{resnet} w/ Cloth label}  & \textcolor{black}{63.08} & \textcolor{black}{75.05} & \textcolor{black}{29.16} & \textcolor{black}{20.89} & \textcolor{black}{31.64} & \textcolor{black}{9.46}\tabularnewline
\hline 
Ours w/o SE & 65.92 & 76.29 & 29.84 & 22.10 & 31.86 & 10.28 \tabularnewline
Ours w/o RN  & 65.51 & 75.89 & 29.37 & 21.29 & 31.67 & 10.05 \tabularnewline
Ours w/o Attn & 64.09 & 76.70 & 28.82 & 22.31 & 34.12 & 9.78 \tabularnewline
Ours w/ single CESD & 64.21 & \textbf{77.51} & 29.46 & 23.73 & 34.30 & 10.19 \tabularnewline
\hline 
Ours & \textbf{66.73} & 77.48 & \textbf{31.29} & \textbf{25.15} & \textbf{34.48} & \textbf{11.67} \tabularnewline
\hline 
\end{tabular}}\caption{\label{tab:Results-ablation} Comparing the contributions of each component in our method on LTCC dataset. `w/ Cloth label' means the model is also trained with clothes labels. `RN' and `Attn' refer to the relation network and the design of self-attention in CESD module, respectively. `w/ single CESD' indicates that we only insert one CESD module after \textit{res4} block. Note that all models of variants are trained only with images of identities who have more than one outfit.}
\end{table}

\begin{figure}
\begin{centering}
\includegraphics[scale=0.35]{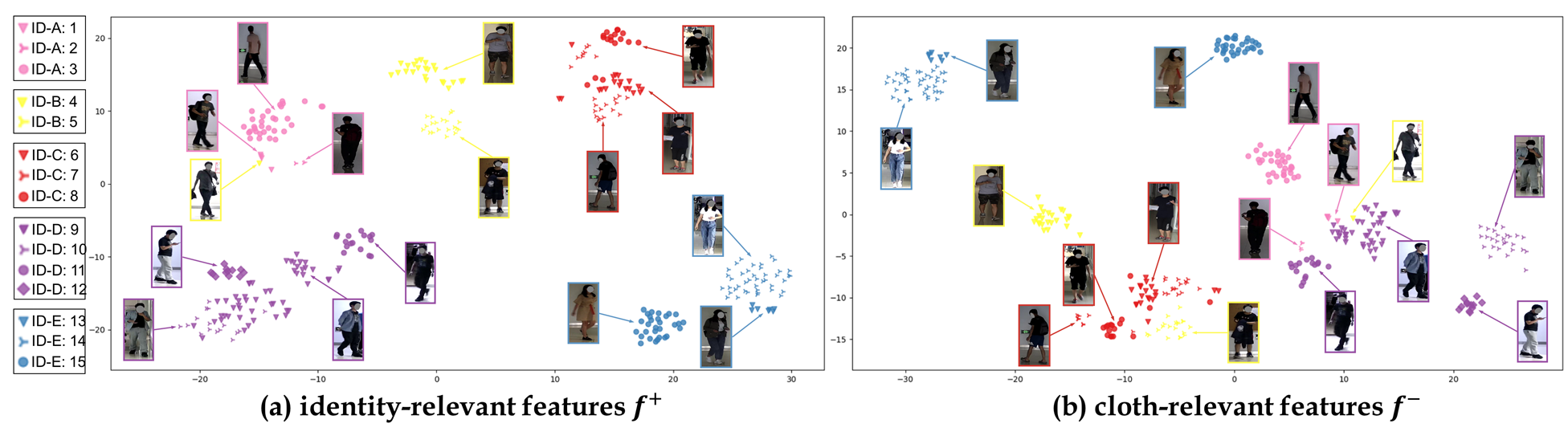} 
\caption{Visualization of the identity-relevant features and cloth-relevant features learned from CESD using t-SNE \cite{maaten2008visualizing}. In (a) and (b), each color represents one identity which is randomly selected from the testing set, and each symbol (circle, rhombus, triangle, cross, \textit{etc.}) with various color indicates different clothes. In the legend, the capital letters behind the `ID' indicate the identity labels, and the numbers are used to represent the clothing category. Best viewed in color and zoomed in.}
\label{fig:tsne} 
\end{centering}
\end{figure}

\noindent \textbf{Effectiveness of shape embedding module.} Our shape embedding (SE) module is designed to exploit the shape information from human joints, so that it can facilitate the task of long-term cloth-changing Re-ID. Here, we compare several variants to evaluate its effectiveness. `Ours w/o SE': a model without the SE module. `Ours w/o RN': a model where the keypoint feature in SE module is encoded directly with several FC layers instead of the relation network. 

We have three observations from the results in Tab.~\ref{tab:Results-ablation}. (1) The variants of `Ours w/o SE' and `Ours w/o RN' achieve very similar results and both are clearly inferior to the full model. It clearly verifies that the shape information can be better encoded with the relations between two different points rather than the feature from an individual point. (2) With the information from human body shape, `Ours' obtains a better performance than `Ours w/o SE' (about $3\%$ higher on Rank-1), which demonstrates the effectiveness of our shape embedding module. (3) Comparing the gap of our model with/without SE module between two different settings, we notice that the shape information contributes about $1\%$ to the accuracy of Rank-1 under the standard setting, but $3$ points under the cloth-changing setting, which shows that under the LTCC ReID task, the biological traits, \textit{e.g.}, body shape, is more important and helpful.

\noindent \textbf{Effectiveness of cloth-elimination shape-distillation module.}
Our proposed CESD module aims to distill the shape features and remove the cloth-relevant features from the input. \textcolor{black}{We consider three variants in the analysis. Concretely, we first compare our full model with `ResNet50 w/ Cloth label', whereby a vanilla ResNet-50 is trained with identity and clothes labels. The results under the two settings in Tab.~\ref{tab:Results-ablation} show that the clothes label helps the model a bit under the standard setting but not on the more challenging cloth-changing setting. Meanwhile, it is clear that it is our model rather than the clothes label that plays the main part.} Secondly, we compare our full model with `Ours w/o Attn', where a variant removes the self-attention block. It shows that `Ours' gets about $2\sim3\%$ higher accuracy on Rank-1/mAP under both settings, which clearly suggests the benefit of the proposed CESD module on LTCC Re-ID by explicitly disentangling and eliminating the cloth-relevant features. Considering the features from the last two blocks of ResNet-50 are deep features, which are more relevant to the specific task, we then evaluate the effect of different number of CESD modules. 
From Tab.~\ref{tab:Results-ablation}, we can conclude that inserting one CESD module after \textit{res3} and \textit{res4} blocks individually is better than `Ours w/ single CESD', which achieves around $2$ points higher on Rank-1 and mAP. It further demonstrates the effectiveness of our CESD module.

\noindent \textbf{Visualization of features learned from CESD.}
As described in Sec.~\ref{sec:method}, our CESD module can disentangle the input feature into two spaces, the identity-relevant and the cloth-relevant. To verify our motivation and better understand its objective, we visualize the learned features from the last CESD module using t-SNE \cite{maaten2008visualizing}. Specifically, we randomly select images of five identities in the testing set, and each of them appeared in more than one outfits. We can observe that (1) with the overview of Fig.~\ref{fig:tsne}, the samples with the same identity are clustered using identity-relevant feature $f^{+}$, and the distances between images which have similar appearances are closer than those with large clothing discrepancy based on the cloth-relevant feature $f^{-}$. (2) As for Fig.~\ref{fig:tsne} (b), images with warm color appearances are clustered at the top of feature space, and those with cold color are centered at the lower right corner. For example, the girl wearing a khaki dress (blue circle) has more similar appearance feature $f^{-}$ to the boy with pink shorts (pink circle), and images at the bottom of the space are clustered due to wearing similar dark clothes. (3) In the identity-relevant feature space of Fig.~\ref{fig:tsne} (a), identities denoted by blue, yellow and purple have smaller intra-class distances comparing with the distances in the cloth-relevant space in (b). (4) Interestingly, there is a special case denoted by a yellow inverted triangle in both spaces. His body shape is similar to the pink one, so in (a) he is aggregated closer to the pink cluster. Meanwhile, he, wearing black pants and dark shorts (affected by illumination), is surrounded by the images having dark clothes in (b). In conclusion, our model can successfully disentangle the identity-relevant features with the cloth-relevant features.

\section{Conclusion}

In this paper, to study the Re-ID problem under more realistic conditions, we focus on Long-Term Cloth-Changing (LTCC) Re-ID, and introduce a new large-scale LTCC Dataset, which has no cloth-consistency constraint. LTCC has 152 identities with 478 different clothes of 17,128 images from 12 cameras, and among them, there are 91 persons showing clothing changes. To further solve the problem of dramatic appearance changes, we propose a task-driven method, which can learn identity-sensitive and cloth-insensitive representations. We utilize the relation between the human keypoints to extract biological structural features and apply attention mechanism to disentangle the identity-relevant features from clothing-related information. The effectiveness of proposed method is validated through extensive experiments.

\noindent \textbf{Acknowledgment.} This work was supported in part by Science and Technology Commission of Shanghai Municipality Projects (19511120700, 19ZR1471800), NSFC Projects (U62076067, U1611461).

\bibliographystyle{splncs04}
\bibliography{egbib}

\begin{thebibliography}{10}
\providecommand{\url}[1]{\texttt{#1}}
\providecommand{\urlprefix}{URL }
\providecommand{\doi}[1]{https://doi.org/#1}

\bibitem{BakCL18}
Bak, S., Carr, P., Lalonde, J.: Domain adaptation through synthesis for
  unsupervised person re-identification. In: European Conference on Computer
  Vision (2018). \doi{10.1007/978-3-030-01261-8\_12},
  \url{https://doi.org/10.1007/978-3-030-01261-8\_12}

\bibitem{barbosa2012re}
Barbosa, I.B., Cristani, M., Del~Bue, A., Bazzani, L., Murino, V.:
  Re-identification with rgb-d sensors. In: European Conference on Computer
  Vision (2012)

\bibitem{berclaz2008behavioral_maps}
Berclaz, J., Fleuret, F., Fua, P.: Multi-camera tracking and atypical motion
  detection with behavioral maps. In: European Conference on Computer Vision
  (2008)

\bibitem{chan2009counting}
Chan, A., Vasconcelos, N.: Bayesian poisson regression for crowd counting. In:
  IEEE International Conference on Computer Vision (2009).
  \doi{10.1109/ICCV.2009.5459191}

\bibitem{chang2018multi}
Chang, X., Hospedales, T.M., Xiang, T.: Multi-level factorisation net for
  person re-identification. In: CVPR. vol.~1, p.~2 (2018)

\bibitem{chao2019gaitset}
Chao, H., He, Y., Zhang, J., Feng, J.: Gaitset: Regarding gait as a set for
  cross-view gait recognition. In: AAAI Conference on Artificial Intelligence
  (2019)

\bibitem{deng2009imagenet}
Deng, J., Dong, W., Socher, R., Li, L.J., Li, K., Fei-Fei, L.: Imagenet: A
  large-scale hierarchical image database. In: IEEE Conference on Computer
  Vision and Pattern Recognition (2009)

\bibitem{fang2017rmpe}
Fang, H.S., Xie, S., Tai, Y.W., Lu, C.: {RMPE}: Regional multi-person pose
  estimation. In: IEEE International Conference on Computer Vision (2017)

\bibitem{ghiasi2017exploring}
Ghiasi, G., Lee, H., Kudlur, M., Dumoulin, V., Shlens, J.: Exploring the
  structure of a real-time, arbitrary neural artistic stylization network.
  arXiv preprint arXiv:1705.06830  (2017)

\bibitem{Gong_2017_CVPR}
Gong, K., Liang, X., Zhang, D., Shen, X., Lin, L.: Look into person:
  Self-supervised structure-sensitive learning and a new benchmark for human
  parsing. In: IEEE Conference on Computer Vision and Pattern Recognition
  (2017)

\bibitem{he2017mask}
He, K., Gkioxari, G., Doll{\'a}r, P., Girshick, R.: Mask r-cnn. In: IEEE
  International Conference on Computer Vision (2017)

\bibitem{resnet}
He, K., Zhang, X., Ren, S., Sun, J.: Deep residual learning for image
  recognition. In: IEEE Conference on Computer Vision and Pattern Recognition
  (2015)

\bibitem{he2019foreground}
He, L., Wang, Y., Liu, W., Zhao, H., Sun, Z., Feng, J.: Foreground-aware
  pyramid reconstruction for alignment-free occluded person re-identification.
  In: IEEE International Conference on Computer Vision (2019)

\bibitem{huang2017arbitrary}
Huang, X., Belongie, S.: Arbitrary style transfer in real-time with adaptive
  instance normalization. In: IEEE International Conference on Computer Vision
  (2017)

\bibitem{huang2019celebrities}
Huang, Y., Wu, Q., Xu, J., Zhong, Y.: Celebrities-reid: A benchmark for clothes
  variation in long-term person re-identification. In: International Joint
  Conference on Neural Networks (2019)

\bibitem{huang2019beyond}
Huang, Y., Xu, J., Wu, Q., Zhong, Y., Zhang, P., Zhang, Z.: Beyond scalar
  neuron: Adopting vector-neuron capsules for long-term person
  re-identification. IEEE Transactions on Circuits and Systems for Video
  Technology  (2019)

\bibitem{joon2015person}
Joon~Oh, S., Benenson, R., Fritz, M., Schiele, B.: Person recognition in
  personal photo collections. In: IEEE International Conference on Computer
  Vision (2015)

\bibitem{crowduser}
Kittur, A., Chi, E.H., Suh., B.: Crowdsourcing user studies with mechanical
  turk. In: ACM Computer-Human Interaction (CHI) Conference on Human Factors in
  Computing Systems (2008)

\bibitem{cuhk01}
Li, W., Zhao, R., X.Wang: Human re-identification with transferred metric
  learning. In: Asian Conference on Computer Vision (2012)

\bibitem{deepreid}
Li, W., Zhao, R., Xiao, T., Wang, X.: Deepreid: Deep filter pairing neural
  network for person re-identification. In: IEEE Conference on Computer Vision
  and Pattern Recognition (2014)

\bibitem{li2018harmonious}
Li, W., Zhu, X., Gong, S.: Harmonious attention network for person
  re-identification. In: IEEE Conference on Computer Vision and Pattern
  Recognition (2018)

\bibitem{liang2018look}
Liang, X., Gong, K., Shen, X., Lin, L.: Look into person: Joint body parsing \&
  pose estimation network and a new benchmark. IEEE Transactions on Pattern
  Analysis and Machine Intelligence  (2018)

\bibitem{XQDA}
Liao, S., Hu, Y., Zhu, X., Li., S.Z.: Person re-identification by local maximal
  occurrence representation and metric learning. In: IEEE Conference on
  Computer Vision and Pattern Recognition (2015)

\bibitem{liu2018iqiyi}
Liu, Y., Shi, P., Peng, B., Yan, H., Zhou, Y., Han, B., Zheng, Y., Lin, C.,
  Jiang, J., Fan, Y., et~al.: iqiyi-vid: A large dataset for multi-modal person
  identification. arXiv preprint arXiv:1811.07548  (2018)

\bibitem{maaten2008visualizing}
Maaten, L.v.d., Hinton, G.: Visualizing data using t-sne. Journal of Machine
  Learning Research  (2008)

\bibitem{miao2019pose}
Miao, J., Wu, Y., Liu, P., Ding, Y., Yang, Y.: Pose-guided feature alignment
  for occluded person re-identification. In: IEEE International Conference on
  Computer Vision (2019)

\bibitem{munaro2014one}
Munaro, M., Fossati, A., Basso, A., Menegatti, E., Van~Gool, L.: One-shot
  person re-identification with a consumer depth camera. In: Person
  Re-Identification (2014)

\bibitem{paisitkriangkrai2015learning}
Paisitkriangkrai, S., Shen, C., van~den Hengel, A.: Learning to rank in person
  re-identification with metric ensembles. In: IEEE Conference on Computer
  Vision and Pattern Recognition (2015)

\bibitem{perez2017learning}
Perez, E., De~Vries, H., Strub, F., Dumoulin, V., Courville, A.: Learning
  visual reasoning without strong priors. arXiv preprint arXiv:1707.03017
  (2017)

\bibitem{perez2018film}
Perez, E., Strub, F., De~Vries, H., Dumoulin, V., Courville, A.: Film: Visual
  reasoning with a general conditioning layer. In: AAAI Conference on
  Artificial Intelligence (2018)

\bibitem{qian2017pose}
Qian, X., Fu, Y., Wang, W., Xiang, T., Wu, Y., Jiang, Y.G., Xue, X.:
  Pose-normalized image generation for person re-identification. European
  Conference on Computer Vision  (2018)

\bibitem{qian2019leader}
Qian, X., Fu, Y., Xiang, T., Jiang, Y.G., Xue, X.: Leader-based multi-scale
  attention deep architecture for person re-identification. IEEE Transactions
  on Pattern Analysis and Machine Intelligence  (2019)

\bibitem{ristani2016MTMC}
Ristani, E., Solera, F., Zou, R., Cucchiara, R., Tomasi, C.: Performance
  measures and a data set for multi-target, multi-camera tracking. In: European
  Conference on Computer Vision Workshop (2016)

\bibitem{santoro2017simple}
Santoro, A., Raposo, D., Barrett, D.G., Malinowski, M., Pascanu, R., Battaglia,
  P., Lillicrap, T.: A simple neural network module for relational reasoning.
  In: Neural Information Processing Systems (2017)

\bibitem{shen2018end}
Shen, Y., Xiao, T., Li, H., Yi, S., Wang, X.: End-to-end deep kronecker-product
  matching for person re-identification. In: IEEE Conference on Computer Vision
  and Pattern Recognition (2018)

\bibitem{sun2019dissecting}
Sun, X., Zheng, L.: Dissecting person re-identification from the viewpoint of
  viewpoint. In: IEEE Conference on Computer Vision and Pattern Recognition
  (2019)

\bibitem{sun2018beyond}
Sun, Y., Zheng, L., Yang, Y., Tian, Q., Wang, S.: Beyond part models: Person
  retrieval with refined part pooling (and a strong convolutional baseline).
  In: European Conference on Computer Vision (2018)

\bibitem{tang2018pyramidbox}
Tang, X., Du, D.K., He, Z., Liu, J.: Pyramidbox: A context-assisted single shot
  face detector. In: European Conference on Computer Vision (2018)

\bibitem{ulyanov2016instance}
Ulyanov, D., Vedaldi, A., Lempitsky, V.: Instance normalization: The missing
  ingredient for fast stylization. arXiv preprint arXiv:1607.08022  (2016)

\bibitem{wang2019color}
Wang, G., Yang, Y., Cheng, J., Wang, J., Hou, Z.: Color-sensitive person
  re-identification. In: International Joint Conference on Artificial
  Intelligence (2019)

\bibitem{wang2018learning}
Wang, G., Yuan, Y., Chen, X., Li, J., Zhou, X.: Learning discriminative
  features with multiple granularities for person re-identification. In: ACM
  International Conference on Multimedia (2018)

\bibitem{wang2008correspondencefree}
Wang, X., Tieu, K., Grimson, W.: Correspondence-free multi-camera activity
  analysis and scene modeling. In: IEEE Conference on Computer Vision and
  Pattern Recognition (2008)

\bibitem{wei2018person}
Wei, L., Zhang, S., Gao, W., Tian, Q.: Person transfer gan to bridge domain gap
  for person re-identification. In: IEEE Conference on Computer Vision and
  Pattern Recognition (2018)

\bibitem{xiu2018poseflow}
Xiu, Y., Li, J., Wang, H., Fang, Y., Lu, C.: {Pose Flow}: Efficient online pose
  tracking. In: British Machine Vision Conference (2018)

\bibitem{xue2018clothing}
Xue, J., Meng, Z., Katipally, K., Wang, H., van Zon, K.: Clothing change aware
  person identification. In: IEEE Conference on Computer Vision and Pattern
  Recognition Workshops (2018)

\bibitem{yang2019person}
Yang, Q., Wu, A., Zheng, W.S.: Person re-identification by contour sketch under
  moderate clothing change. IEEE Transactions on Pattern Analysis and Machine
  Intelligence  (2019)

\bibitem{yu2017devil}
Yu, Q., Chang, X., Song, Y.Z., Xiang, T., Hospedales, T.M.: The devil is in the
  middle: Exploiting mid-level representations for cross-domain instance
  matching. arXiv preprint arXiv:1711.08106  (2017)

\bibitem{NullReid}
Zhang, L., Xiang, T., Gong, S.: Learning a discriminative null space for person
  re-identificatio. In: IEEE Conference on Computer Vision and Pattern
  Recognition (2016)

\bibitem{zhang2015beyond}
Zhang, N., Paluri, M., Taigman, Y., Fergus, R., Bourdev, L.: Beyond frontal
  faces: Improving person recognition using multiple cues. In: IEEE Conference
  on Computer Vision and Pattern Recognition (2015)

\bibitem{market1501}
Zheng, L., Shen, L., Tian, L., S.Wang, J.Wang, Tian, Q.: Scalable person
  re-identification: A benchmark. In: IEEE International Conference on Computer
  Vision (2015)

\bibitem{reid_in_wild}
Zheng, L., Zhang, H., Sun, S., Chandraker, M., Tian, Q.: Person
  re-identification in the wild. arXiv preprint arXiv:1604.02531  (2016)

\bibitem{zheng2015partial}
Zheng, W.S., Li, X., Xiang, T., Liao, S., Lai, J., Gong, S.: Partial person
  re-identification. In: IEEE International Conference on Computer Vision
  (2015)

\bibitem{zheng2019joint}
Zheng, Z., Yang, X., Yu, Z., Zheng, L., Yang, Y., Kautz, J.: Joint
  discriminative and generative learning for person re-identification. In: IEEE
  Conference on Computer Vision and Pattern Recognition (2019)

\bibitem{zheng2017discriminatively}
Zheng, Z., Zheng, L., Yang, Y.: A discriminatively learned cnn embedding for
  person reidentification. ACM Transactions on Multimedia Computing,
  Communications, and Applications (TOMM)  \textbf{14}(1), ~13 (2017)

\bibitem{zheng2017unlabeled}
Zheng, Z., Zheng, L., Yang, Y.: Unlabeled samples generated by gan improve the
  person re-identification baseline in vitro. In: IEEE International Conference
  on Computer Vision (2017)

\bibitem{zhong2017random}
Zhong, Z., Zheng, L., Kang, G., Li, S., Yang, Y.: Random erasing data
  augmentation. arXiv preprint arXiv:1708.04896  (2017)

\bibitem{zhou2019omni}
Zhou, K., Yang, Y., Cavallaro, A., Xiang, T.: Omni-scale feature learning for
  person re-identification. In: IEEE International Conference on Computer
  Vision (2019)

\end{thebibliography}

\newpage

\appendix
\renewcommand{\appendixname}{Appendix~\Alph{section}}
\section{More details on BIWI dataset}

\iffalse
Due to space limitation, in the main paper, we just reported results on our new LTCC dataset. Here, additional experimental results on an existing, much smaller cloth-changing re-ID dataset are presented and discussed. 

\begin{table}
\centering{}%
\setlength{\tabcolsep}{3mm}{
\begin{tabular}{@{\extracolsep{\fill}}l|cc|cc}
\hline 
\multicolumn{1}{c|}{\multirow{2}{*}{Methods}} & \multicolumn{2}{c|}{Still Setting} & \multicolumn{2}{c}{Walking Setting}\tabularnewline
\cline{2-5}
 & Rank-1 & Rank-5 & Rank-1 & Rank-5 \tabularnewline
\hline
ResNet-50 \cite{resnet} & 41.26 & 65.13 & 37.08 & 65.01 \tabularnewline
PCB \cite{sun2018beyond} & 46.36 & 72.37 & 37.89 & \textbf{70.21} \tabularnewline
\hline 
SPT+ASE$^*$ \cite{yang2019person} & 21.31 & 66.10 & 18.66 & 63.88 \tabularnewline
HACNN \cite{li2018harmonious} & 42.71 & 64.89 & 37.32 & 64.27 \tabularnewline
MuDeep \cite{qian2019leader} & 43.46 & \textbf{73.80} & 38.41 & 65.35 \tabularnewline
OSNet \cite{zhou2019omni} & 47.14 & 72.52 & 38.87 & 61.22 \tabularnewline
\hline 
Ours & \textbf{49.51} & 71.44 & \textbf{39.80} & 66.23 \tabularnewline
\hline 
\end{tabular}}
\caption{\label{tab:Results-DIWI} Results on BIWI dataset. 
`$*$' denotes the results are reported from original paper~\cite{yang2019person}.
Note that, the same baselines (\textit{i.e.}, ResNet-50 \cite{resnet}, PCB~\cite{sun2018beyond}) implemented by us yield better results  than those reported in~\cite{yang2019person}.}
\end{table}
\fi

Due to space limitation, in the main paper, we only reported results on BIWI dataset. Here, we discuss more about this existing and smaller cloth-changing re-ID dataset and the implementation details. 

\noindent\textbf{BIWI dataset.}
BIWI \cite{munaro2014one} is a small-scale person re-identification dataset with cloth changes.
It contains $50$ identities, $28$ of where appeared in two outfits.
It is collected from two cameras  and people are asked to perform a certain routine of motions in front of the cameras such as, 
a rotation around the vertical axis, several head movements (\textit{i.e.}, the action of `still') or 
% two 
walking towards the camera (\textit{i.e.}, the action of `walk'). 
The dataset consists of three subsets: 
`Train' subset with images of all $50$ identities from one camera, 
`Still' and `Walking' subsets of both with images of $28$ cloth-changing people with above two actions from another camera.
% corresponding to the two actions. 
Some examples are shown in Fig.~\ref{fig:dataset_others}(d).

\noindent\textbf{Implementation details.} 
%To further evaluate the efficacy of our method, we conduct experiments on this dataset. Specifically, 
We randomly select $14$ cloth-changing identities as training and the rest of $14$ for testing. 
Considering the small-scale nature of BIWI, we pretrain all models on the LTCC dataset, and then fine-tune them on the training set of BIWI for $80$ epochs with initial learning rate of $0.001$ and decay rate of $0.1$ for every $30$ epochs. 
At the testing stage, we set the test images from `Train' subset as query images, and those from `Still' or `Walking' subset as gallery images. Therefore, our experiment includes two test settings: Still and Walking setting. 
The evaluation metrics used in \cite{deepreid,cuhk01} are adopt to calculate the scores of Rank-$k$ under the multi-shot setting.

\iffalse
\noindent\textbf{Results.} 
From the results shown in Table \ref{tab:Results-DIWI}, we can make the following observations:
(1) Our method achieves the highest performance on Rank-$1$ at both settings (Rank-1 of 49.51\% and 39.80\% compared to the closest competitor OSNet~\cite{zhou2019omni} which gives 47.14\% and 38.87\% at Still and Walking setting respectively).
(2) Our method beats the state-of-the-art competitor  SPT+ASE~\cite{yang2019person}, which is designed for cloth-changing re-ID specifically, by over 28\% and 21\% on Rank-$1$ under the Still and Walking setting, respectively.
Note that we adopt the same training/test split and pretrain strategy as SPT+ASE~\cite{yang2019person}.
(3) We also notice that the performance of Still setting are generally higher than the Walking setting.
This confirms the conclusion drawn in the main paper as  
% It is consistent with our common sense because the images from 
`Still' subset involve fewer variations of pose and occlusions.
% (3) Some simple networks, such as ResNet-50 and PCB, also can get comparative performances, we explain that although BIWI considers the factor of cloth-changing, it has images with higher resolution and fewer challenges of illumination, occlusion, pose and so on. 
\fi

\section{Results on Celeb-reID dataset}
We further carry out new experiments on Celeb-reID \cite{huang2019celebrities,huang2019beyond}, and compare with several strong competitors, including MGN \cite{wang2018learning}, ReIDCaps \cite{huang2019beyond} and HACNN \cite{li2018harmonious}.

\noindent\textbf{Celeb-reID dataset.} Celebrities-reID dataset \cite{huang2019celebrities,huang2019beyond} contains $1052$ IDs with $34,186$ images. The images are collected by the street snap-shots of celebrities on Internet. Considering the same person usually does not wear the same clothing twice in snap-shots, this dataset is also suitable for the clothes variation person re-ID study. Specifically, more than 70\% of the images of each person show different clothes on average. Some examples are shown in Fig.~\ref{fig:dataset_others}(e).

\noindent\textbf{Implementation details.} Following the standard training and testing setting in \cite{huang2019beyond}, we use $632$ identities as training set and $420$ identities for testing. Besides, we set a random unique number for each training image as the cloth label since it doesn't contain any clothes annotation, and replace ResNet-50 backbone with DenseNet121 for a fair comparison.

\noindent\textbf{Results.} As shown in Table~\ref{tab:Results-celeb}, our model surprisingly gets Rank1/mAP 50.9\%/9.8\% with the dummy cloth label, while PCB 37.1\%/8.2\%, HACNN 47.6\%/9.5\% and ReIDCaps 51.2\%/9.8\%. It clearly suggests the effectiveness of our model.

\begin{table}
\centering{}%
\setlength{\tabcolsep}{3mm}{
\begin{tabular}{@{\extracolsep{\fill}}l|ccc}
\hline 
\multicolumn{1}{c|}{Methods} & Rank-1 & Rank-5 & mAP\tabularnewline
\hline
IDE+ (DenseNet121) \cite{reid_in_wild} & 42.9 & 56.4 & 5.9\tabularnewline
PCB \cite{sun2018beyond} & 37.1 & 57.0 & 8.2\tabularnewline
HACNN \cite{li2018harmonious} & 47.6 & 63.3 & 9.5\tabularnewline
ResNet-Mid \cite{yu2017devil} & 43.3 & 54.6 & 5.8 \tabularnewline
Two-Stream \cite{zheng2017discriminatively} & 36.3 & 54.5 & 7.8\tabularnewline
MLFN \cite{chang2018multi} & 41.4 & 54.7 & 6.0\tabularnewline
MGN \cite{wang2018learning} & 49.0 & 64.9 & \textbf{10.8}\tabularnewline
ReIDCaps$^*$ \cite{huang2019beyond} & \textbf{51.2} & 65.4 & 9.8\tabularnewline
\hline 
Ours & 50.9 & \textbf{66.3} & 9.8\tabularnewline
\hline 
\end{tabular}}
\caption{\label{tab:Results-celeb} Results on Celeb-reID dataset. 
`$*$' denotes the results are reported under the setting of `without human body parts partition'.}
\end{table}

\section{Evaluations on model generalization}

Intuitively, the body shape information should be more robust to the domain gap than other appearance information, \textit{e.g.}, clothing, since the distribution of appearance can be easily changed by illumination or camera viewing conditions. Here, we conduct experiments by directly applying the best model on LTCC dataset to Market-1501 and DukeMTMC-reID without any fine-tuning. 

The results are listed in Tab. \ref{tab:Results-generalization}, and our model significantly outperforms other approaches.
For baseline `ResNet-50' and `PCB', `Ours' outperforms them with a large margin of more than $6\%$ accuracy improvement of Rank-1 in both settings.
To be notice, our model beats all of other methods, considering MuDeep and OSNet both can achieve good performance under cross-domain task, which further indicates that our method has strong generalization ability. 

\iffalse
We further evaluate the generalization ability of models on DukeMTMC-reID dataset, which is more challenging than Market-1501 used in the main paper for the same evaluation (samples of both datasets can be found in Fig.~\ref{fig:dataset_others}(a) and (b)). 
As illustrated in Table \ref{tab:Results-generalization}, 
our model beats the recent state-of-the-arts 
HACNN \cite{li2018harmonious}, MuDeep \cite{qian2019leader} and OSNet \cite{zhou2019omni}
by  clear margins
and obtains  superior performance on all metrics.
% two baselines ResNet-50 \cite{resnet}, PCB~\cite{sun2018beyond}, with huge margin, nearly $10\%$ Rank-1 gap. 
% Furthermore, comparing with several powerful Re-ID methods, \textit{i.e.}, HACNN \cite{li2018harmonious}, MuDeep \cite{qian2019leader} and OSNet \cite{zhou2019omni}, our approach can obtain the superior performance on all metrics.
We can see the same trend on both Market-1501 and DukeMTMC-reID datasets.
This suggests that by modeling the structural information,
our method attains stronger ability for generalization.
\fi

\begin{table*}
\centering{}%
\setlength{\tabcolsep}{0.5mm}{
\begin{tabular}{@{\extracolsep{\fill}}l|cc|cc|cc|cc}
\hline 
\multicolumn{1}{c|}{\multirow{3}{*}{Methods}} & \multicolumn{4}{c|}{LTCC $\to$ Market-1501 \cite{market1501}} & \multicolumn{4}{c}{LTCC $\to$ DukeMTMC-reID \cite{zheng2017unlabeled}}\tabularnewline
\cline{2-9}
& \multicolumn{2}{c|}{Standard Setting} & \multicolumn{2}{c|}{Standard Setting$^\dag$} & \multicolumn{2}{c|}{Standard Setting} & \multicolumn{2}{c}{Standard Setting$^\dag$}\tabularnewline
\cline{2-9} 
 & Rank-1 & mAP & Rank-1& mAP & Rank-1 & mAP & Rank-1& mAP\tabularnewline
\hline 
ResNet-50 \cite{resnet} & 24.70 & 9.57 & 22.06& 7.89 & 16.87 & 6.26 & 13.64 & 4.90 \tabularnewline
PCB \cite{sun2018beyond} & 31.15 & 13.35 & 28.47 & 11.32 & 18.22 & 7.91 & 14.72 & 6.03 \tabularnewline
\hline
HACNN \cite{li2018harmonious} & 26.90 & 10.35 & 23.18 & 8.29 & 17.35 & 6.82 & 13.94 & 5.76 \tabularnewline
MuDeep \cite{qian2019leader} & 29.36 & 11.22 & 22.27 & 8.21 & 18.53 & 7.65 & 14.27 & 5.89 \tabularnewline
OSNet \cite{zhou2019omni} & 34.33& 15.59 & 32.77  & 14.01 & 13.15 & 8.92 & 21.49 & 7.87 \tabularnewline
\hline
Ours & \textbf{37.38} & \textbf{16.97} & \textbf{34.44}& \textbf{14.09} & \textbf{24.15} & \textbf{9.80} & \textbf{23.47} & \textbf{9.47} \tabularnewline
\hline 
\end{tabular}}
\vspace{2mm}
\caption{\label{tab:Results-generalization} Results of the models under cross-domain Re-ID, that is,  trained on LTCC dataset and evaluated directly on Market-1501 and DukeMTMC-reID. `$\dag$' denotes that only the images with cloth changes are used for training.}
\end{table*}

\begin{figure*}
\begin{centering}
\includegraphics[scale=0.23]{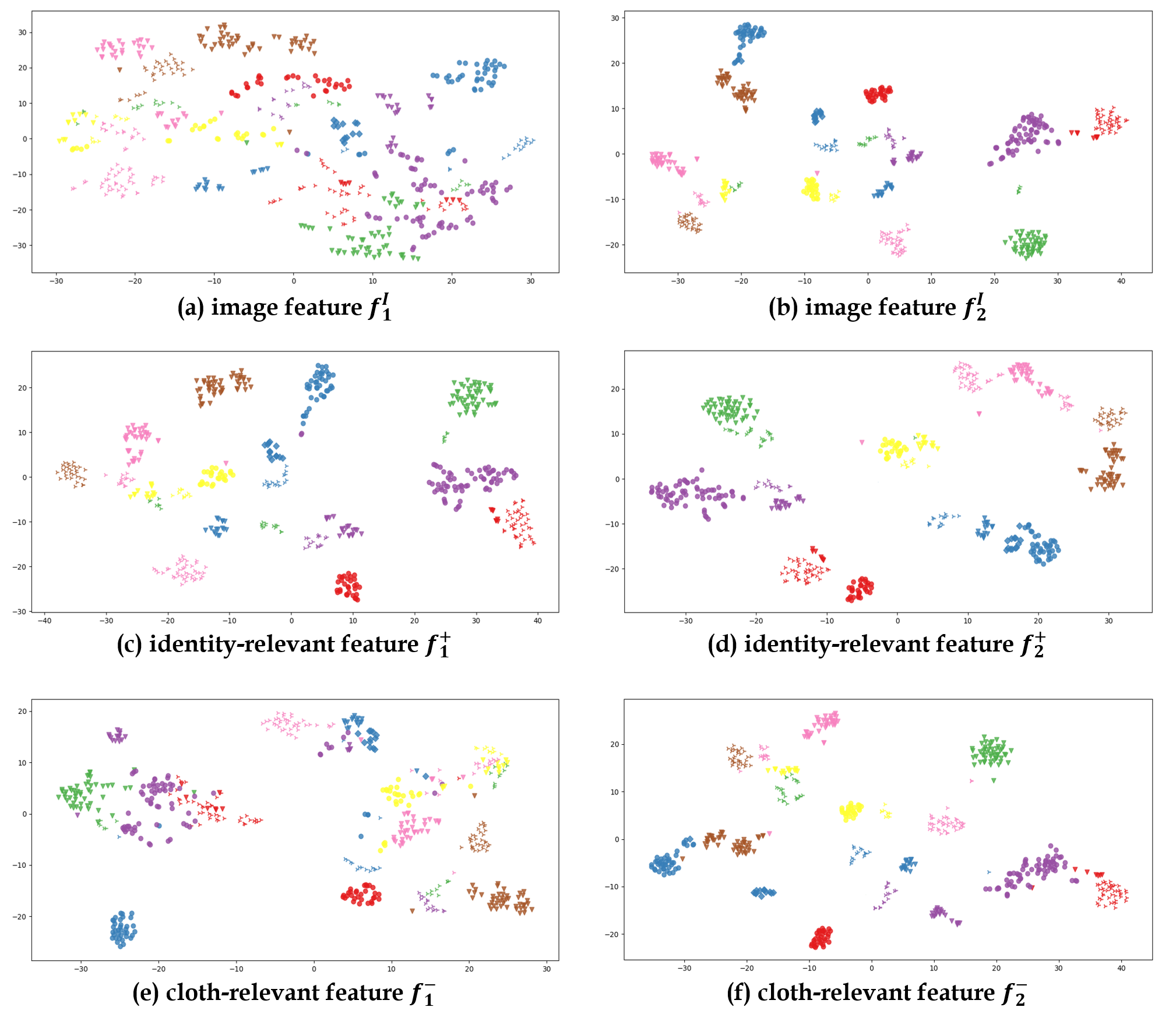} 
\par\end{centering}
\caption{Visualization of six different features extracted from our model using t-SNE \cite{maaten2008visualizing}. The $f_{1}^{I}$, $f_{1}^{+}$ and $f_{1}^{-}$ come from the first CESD module, and the $f_{2}^{I}$, $f_{2}^{+}$ and $f_{2}^{-}$ are generated from the second CESD module. Each color represents one identity which is randomly selected from the testing set, and each symbol (circle, rhombus, triangle, \textit{etc.}) with various color indicates different clothes. Best viewed in color and zoom in. \label{fig:tsne-six}}
\end{figure*}

\section{More visualizations and discussions}

\noindent \textbf{Visualization of features learned from CESD.} 
Due to space limitation, we only show the distribution of two disentangled features, the identity-relevant feature $f^{+}$ and the cloth-relevant feature $f^{-}$ from the last CESD module in the main paper. 
Here, we further visualize all the features associated with two CESD modules in our model to comprehensively analyze the importance and effectiveness of our proposed CESD module.
Specially, we denote the input feature, two outputs of the identity-relevant feature and the cloth-relevant feature in the first CESD module (followed by \textit{res3} block) as $f^{I}_{1}$, $f^{+}_{1}$ and $f^{-}_{1}$, respectively.
Similarly, the associated features in the second CESD module (followed by \textit{res4} block) are defined as $f^{I}_{2}$, $f^{+}_{2}$ and $f^{-}_{2}$. 
As shown in Fig. \ref{fig:tsne-six}, we add more identities in the testing set for visualization, each color represents one identity, and each symbol (\emph{e.g.},~circle, rhombus, triangle, cross) with various color indicates different clothes.
From Fig. \ref{fig:tsne-six}, we can make the following observations:

First, the distribution of features in (a) is relatively more chaotic. 
After going through the first CESD module, it becomes ordered and have the rudiment of clustering, as shown in (c) and (b). 
% Progressively, 
The final output features in (d) are clearly aggregated according to the identity information. 
It strongly indicates the effectiveness of our proposed CESD module in tackling the problem of LTCC Re-ID.

Secondly, from top to bottom (\textit{i.e.}, (a)-(c)-(e) and (b)-(d)-(f)), we observe that our proposed CESD module is able to disentangle the cloth-relevant feature from the input image feature. 
In the space of the identity-relevant feature $f^{+}_{i}$, different symbols with the same color are grouped together based on the information of identities. 
On the contrary, the images with similar clothes are clustered regardless of identities in the space of the cloth-relevant feature $f^{-}_{i}$.

Lastly, from left to right (\textit{i.e.}, (a)-(b), (c)-(d) and (e)-(f)), it is shown that applying two CESD modules can better remove the cloth-sensitive features and distill more discriminative features for person Re-ID. It is not only because the deeper features are more relevant to the specific task, but also because our proposed CESD module plays a key role of cloth-elimination and shape-distillation.

\begin{figure}
\begin{centering}
\includegraphics[scale=0.7]{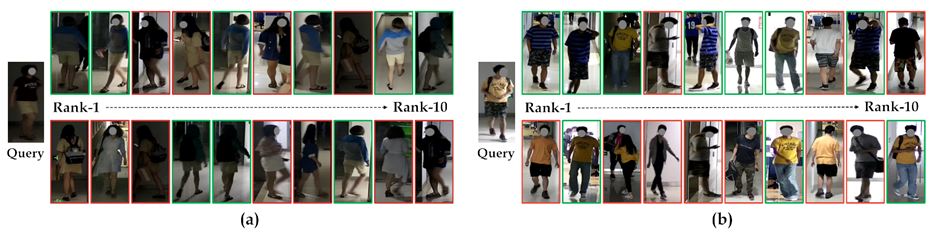} 
\caption{Visualization of retrieval results generated by `Ours' (upper) and `ResNet-50' (lower) under the cloth-changing setting. The images with green borders are the correct results, those with red border are wrong. Best viewed in color and zoomed in.}
\label{fig:Rank10} 
\end{centering}
\end{figure}

\begin{figure*}
\begin{centering}
\includegraphics[scale=0.3]{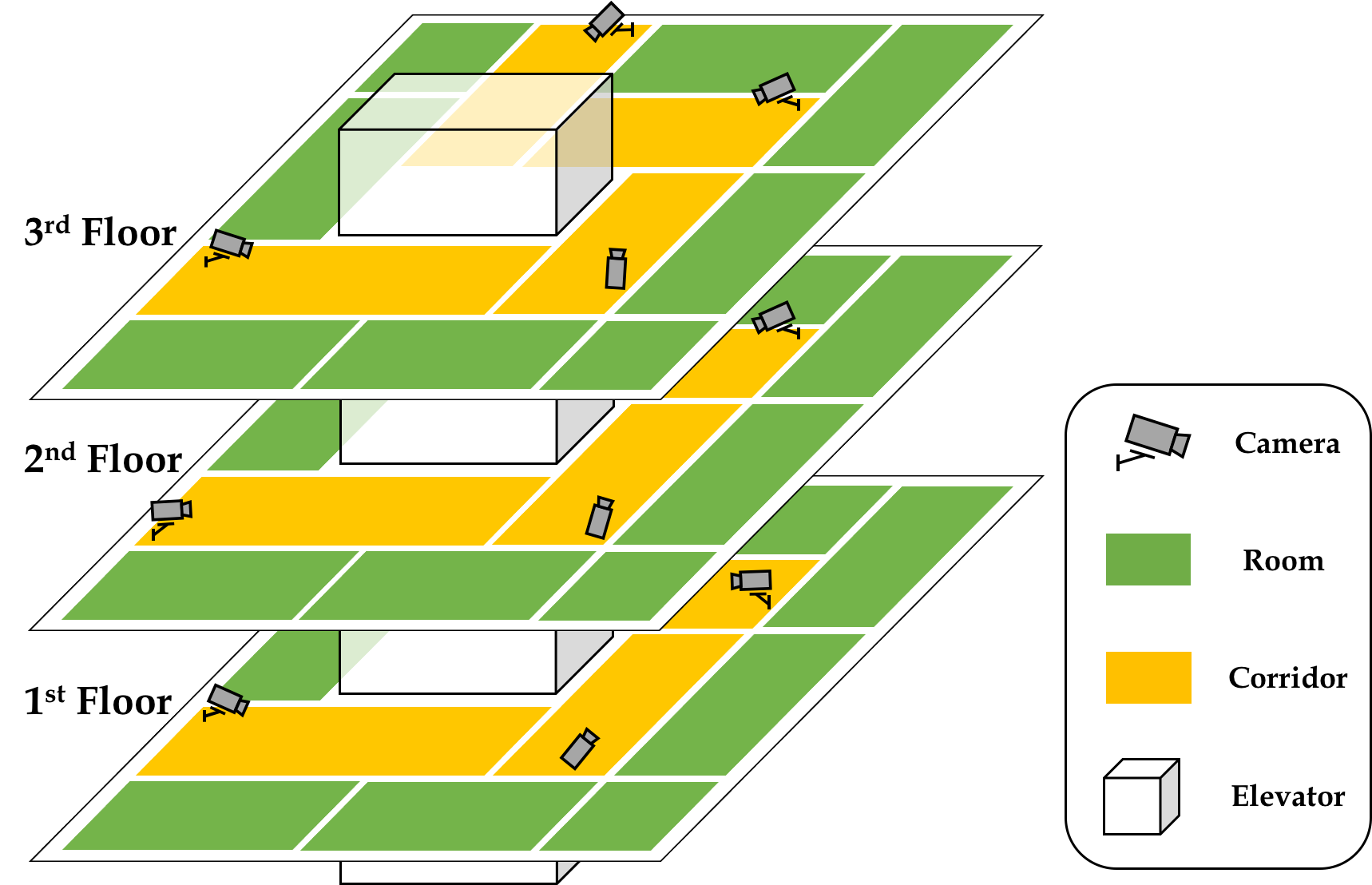} 
\par\end{centering}
\caption{The plan of the camera layout for collecting LTCC data.
\label{fig:layout}}
\end{figure*}

\noindent \textbf{Visualization of retrieval samples.} 
To intuitively demonstrate the ability of our proposed framework in addressing the task of LTCC Re-ID, given same query images, we visualize the top 10 ranked retrieval results of our full model and ResNet-50 under the cloth-changing setting in Fig.~\ref{fig:Rank10}. We can discover that our proposed model can better match the required images with different clothes. For example, the top-2 images in Fig.~\ref{fig:Rank10} (b), which are correctly retrieved by our model, are dressed in different clothes. On the contrary, the ResNet-50 model pays more attention to the similar color and type of yellow shorts. Interestingly, we notice from Fig.~\ref{fig:Rank10} (a) that on account of task-driven training, the ResNet-50 model also can learn some cloth-insensitive features. However, comparing the top-10 results, it cannot tackle the problem of LTCC Re-ID problem well. As a result, these clearly demonstrate that our method devotes more to the shape information rather than the appearance, so it can solve the dramatic changes of appearance caused by cloth changes to some extent.

\begin{figure*}
\begin{centering}
\includegraphics[scale=0.55]{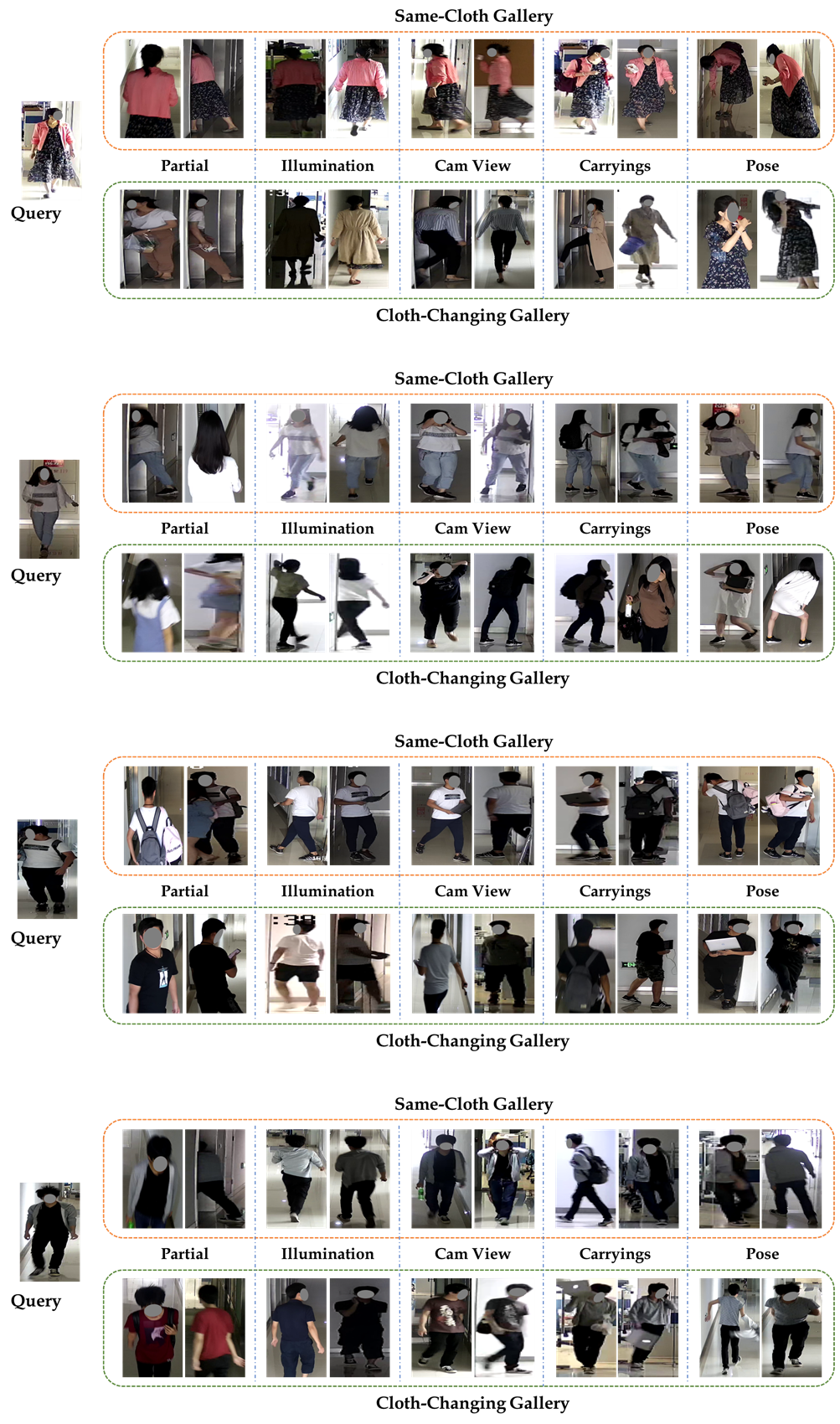} 
\par\end{centering}
\caption{Samples from LTCC dataset. Given the queries, we show the corresponding gallery images with the same clothing, and five different outfits under other variations. 
\label{fig:LTCC}}
\end{figure*}

\begin{figure*}
\begin{centering}
\includegraphics[scale=0.25]{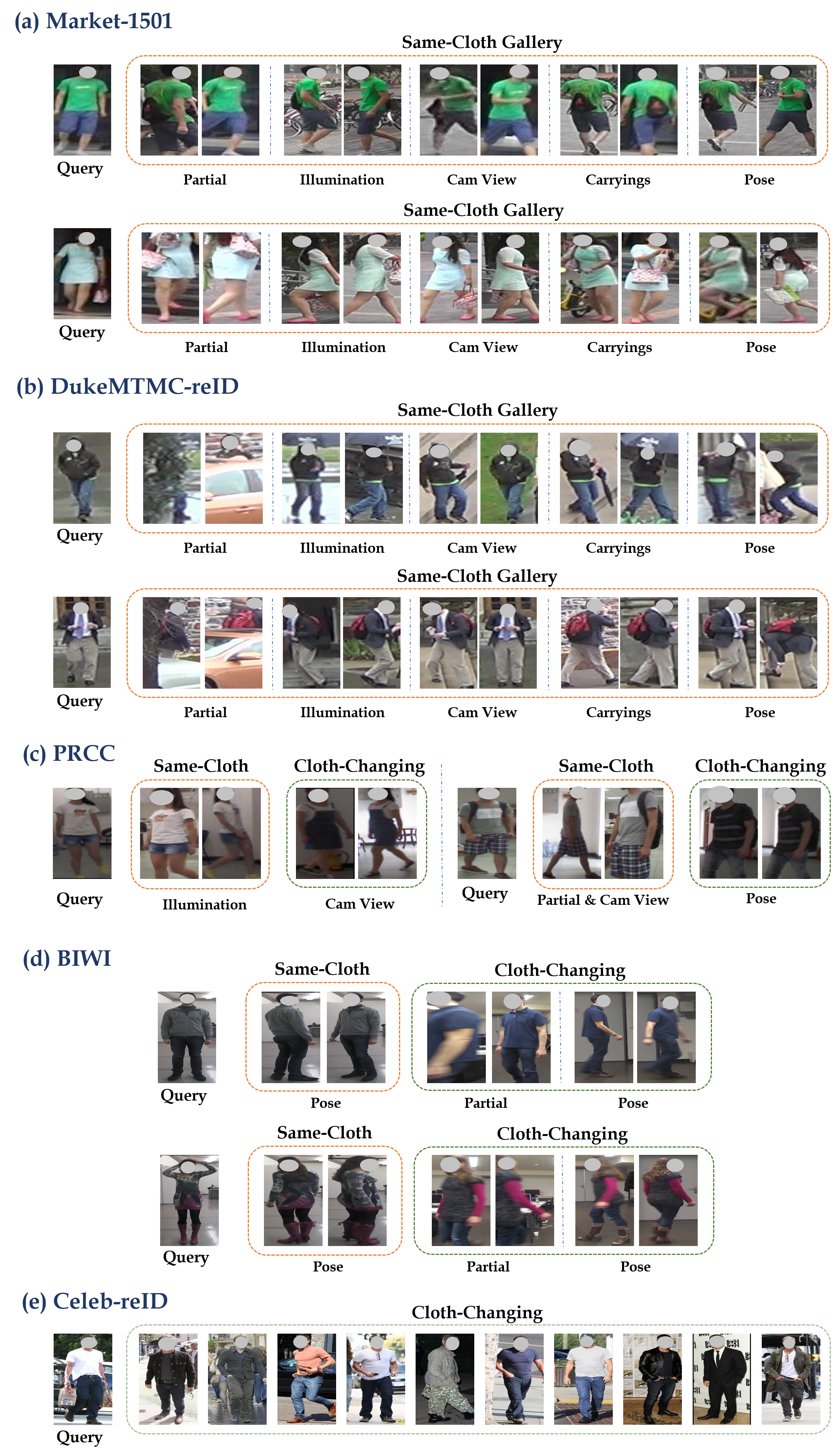} 
\par\end{centering}
\caption{Samples from Market-1501, DukeMTMC-reID, PRCC, BIWI and Celeb-reID datasets. 
\label{fig:dataset_others}}
\end{figure*}

\section{More details on LTCC dataset}

\noindent \textbf{More details on data collection. }
To collect uniform and diverse data for long-term cloth-changing Re-ID, we utilize an existing CCTV system to record videos in 24 hours a day over two months. This CCTV system contains twelve cameras installed on three floors in an office building, shown in Fig. \ref{fig:layout}.
After carefully annotation work, we release the first version of LTCC dataset, which contains $17,138$ person images of $152$ identities with $478$ outfits. More identities will be followed in the future.

\noindent \textbf{Comparison with Previous Datasets. } As shown in Fig. \ref{fig:LTCC}, our proposed LTCC dataset contains cloth-changing identities with huge occlusion, illumination, camera view, carrying and pose variations, which is more realistic for the study of long-term person re-identification. We also compare our LTCC dataset with several widely-used STCC Re-ID datasets (Market-1501 \cite{market1501} and DukeMTMC-reID \cite{zheng2017unlabeled}) and two related cloth-changing Re-ID datasets (BIWI \cite{munaro2014one}, Celeb-reID \cite{huang2019celebrities,huang2019beyond} and PRCC \cite{yang2019person}) in Fig.~\ref{fig:dataset_others}.

Specifically, (1) comparing with Market-1501 and DukeMTMC-reID, these two general STCC Re-ID datasets are limited in clothing, carrying, illumination and other variations for each person. (2) The BIWI dataset, which only contains $28$ people with two different clothes,  is collected under several strict constraints, making it nearly have no variations of view-angle, occlusion, carrying or illumination. (3) The Celeb-reID dataset contains the images from the street snap-shots of celebrities crawled on Internet. And considering that people are usually in the front view from the snap-shots, there only a few back view images are included in the collection. (4) Comparing with PRCC dataset\footnote{PRCC dataset is not yet available at the time of submission, so samples in Fig.~\ref{fig:dataset_others}(c) are directly obtained from \cite{yang2019person} rather than from the actual dataset.}, it contains less drastic clothing changes and bare hairstyle changes. Furthermore, with only $3$ cameras instead of $12$ in our LTCC dataset, it is limited in view-angle, carrying and illumination changes. 

\end{document}